%% file: main.tex
\documentclass[sigconf, nonacm]{acmart}
\AtBeginDocument{%
  }

\usepackage{xspace}
\usepackage{xcolor}
\usepackage{multirow}
\usepackage{colortbl}
\usepackage{bbm}
\usepackage{bbding}
\definecolor{mygray}{gray}{.9}
\definecolor{mypink}{rgb}{.99,.91,.95}
\definecolor{mycyan}{cmyk}{.3,0,0,0}
\definecolor{CQColor}{rgb}{0.0,0.0,1.0} 

\definecolor{CQRColor}{rgb}{1.0,0.0,1.0} 

\definecolor{CQXXYColor}{rgb}{1.0,0.0,0.0} 

\definecolor{emphasizeColor}{HTML}{2755FF}
\newcommand{\emphasize}[1]{{\color{emphasizeColor}#1}}
\definecolor{baseColor}{rgb}{0.75,0.05,0.1}
\newcommand{\base}[1]{{\color{baseColor}#1}}
\definecolor{checkmarkColor}{rgb}{0.1,0.75,0.1}
\newcommand{\checkc}[1]{{\color{checkmarkColor}#1}}
\definecolor{demphcolor}{RGB}{144,144,144}

\definecolor{mygray}{gray}{0.4}
\newlength\savewidth

\newcommand{\tocite}[1]{\textcolor{red}{[TO CITE]}}

\begin{document}
\title{Logic Diffusion for Knowledge Graph Reasoning}

\author{Xiaoying Xie}
\authornote{Both authors contributed equally to this research.}
\email{xiaoyingx@stu.xjtu.edu.cn}
\affiliation{%
  \institution{Xi'an Jiaotong University}
  \country{}
}
\author{Biao Gong}
\authornotemark[1]
\email{a.biao.gong@gmail.com}

\author{Yiliang Lv}
\email{401851090@qq.com}
\affiliation{%
  \country{}
}

\author{Zhen Han}
\email{hanzhn@qq.com}
\affiliation{%
  \institution{Xi'an Jiaotong University}
  \country{}
}

\author{Guoshuai Zhao}
\email{guoshuai.zhao@xjtu.edu.cn}
\affiliation{%
  \institution{Xi'an Jiaotong University}
  \country{}
}

\author{Xueming Qian}
\email{qianxm@mail.xjtu.edu.cn}
\affiliation{%
  \institution{Xi'an Jiaotong University}
  \country{}
}
\input{sections/0_abs.tex}
\maketitle

\input{sections/1_intro.tex}
\input{sections/2_rela.tex}
\input{sections/3_pre.tex}
\input{sections/4_method.tex}

\input{sections/5_exp.tex}

\input{sections/6_con.tex}
\input{sections/7_ref.tex}
\end{document}

%% file: sections/0_abs.tex
\begin{abstract}
Most recent works focus on answering first order logical queries to explore the knowledge graph reasoning via multi-hop logic predictions.
However, existing reasoning models are limited by the circumscribed logical paradigms of training samples, which leads to a weak generalization of unseen logic.
To address these issues, we propose a plug-in module called \textit{Logic Diffusion (LoD)} to discover unseen queries from surroundings and achieves dynamical equilibrium between different kinds of patterns.
The basic idea of \textit{LoD} is relation diffusion and sampling sub-logic by random walking as well as a special training mechanism called gradient adaption.
%
Besides, \textit{LoD} is accompanied by a novel loss function to further achieve the robust logical diffusion when facing noisy data in training or testing sets.
%
Extensive experiments on four public datasets demonstrate the superiority of mainstream knowledge graph reasoning models with \textit{LoD} over state-of-the-art.
Moreover, our ablation study proves the general effectiveness of \textit{LoD} on the noise-rich knowledge graph.
\end{abstract}

%% file: sections/1_intro.tex
\section{Introduction}
Knowledge graph (KG) provides a structural data representation which is organized as triples of entity pairs and relationships \cite{TransE&FB15k&WN18,NELL995, FB15k-237}. In recent year, KG-based common sense reasoning algorithms usually combine mathematical logic \cite{FOL1,FOL2}, relational path \cite{MINERVA,PRA,dapath}, distributional representation \cite{ComplEx,RotatE,JointE}, etc. \cite{CQD,lego,line} with deep learning models to answer First-Order-Logical (FOL) queries, which greatly enhanced reasoning performance and achieved generalization. Different from the basic structural triplets like $ (e_s, r, e_o) $, FOL implements logic by existential quantification ($ \exists$), conjunction ($\wedge$), disjunction ($\vee$), and negation ($\neg$), which is well suited for describing relationships. Around such logical units, the learning objective of models usually focuses on logical mappings rather than representations to achieve the best logical answers.
However, one of the limitations of this former is that the logical paradigms need to be defined manually (e.g., \cite{fbeate1,zhang2021fact}). Also, as mentioned above, since the weak ability of learning representations, the impact of unseen FOL and noisy data with unreliable logic on reasoning performance is catastrophic. Coincidentally, these two elements are abundant and inevitable in the real-world knowledge graph \cite{liu2021neural,Kg-fid,han2020open}. 

\begin{figure}
	\begin{center}
 	\includegraphics[width=0.9\linewidth]{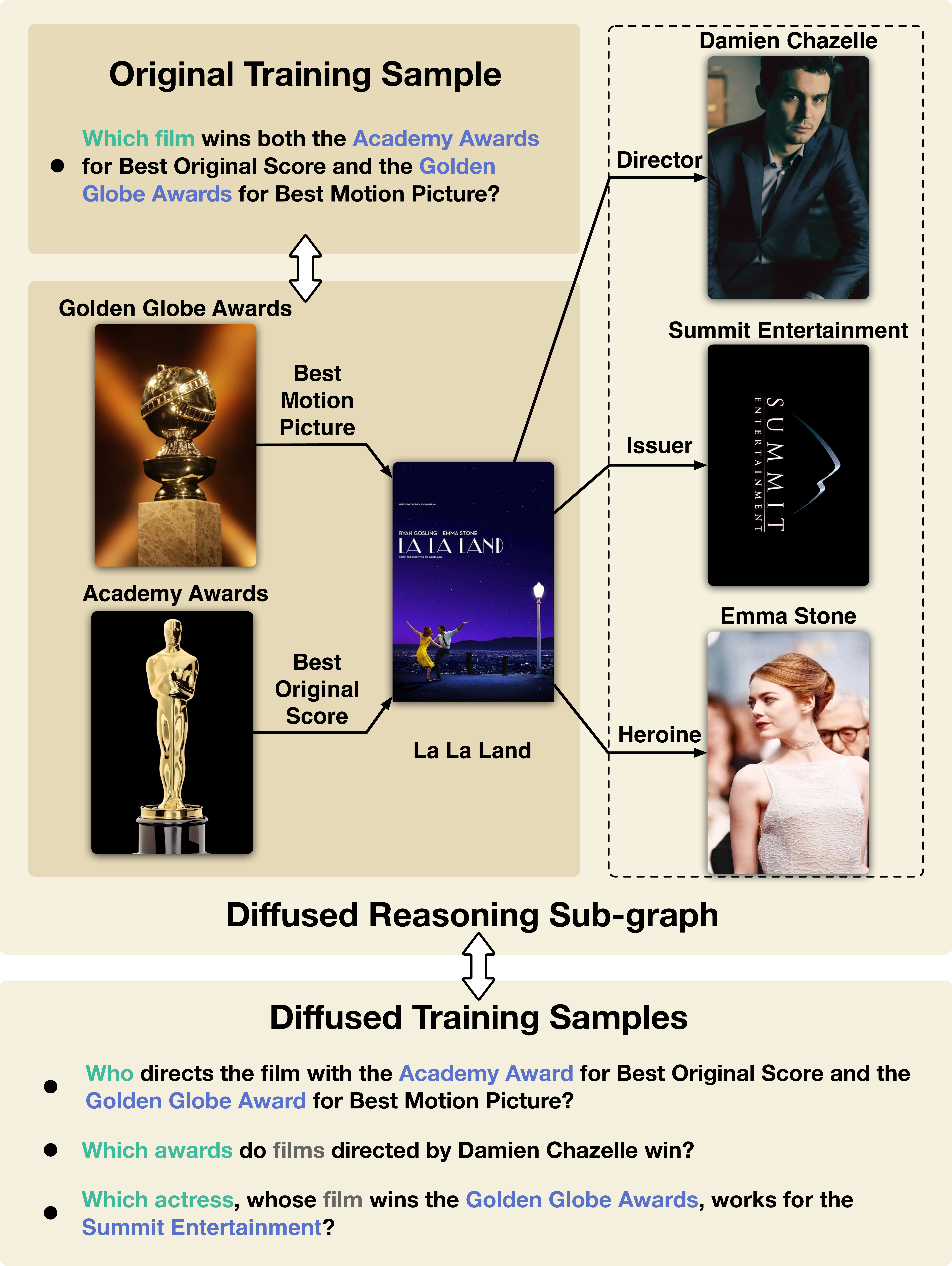}
	\end{center}
	\caption{Illustration of original training samples and extended training samples after the diffusion. How to discover unseen paradigms from seen ones is what \textit{LoD} tries to solve.}
	\label{fig:1}
\end{figure}

Based on the above thinking, we summarize two major problems in the real-world knowledge graph: (1) discovering unseen FOL paradigms, and (2) learning non-duality logic.
Specifically, the unseen FOL refers to the FOL that does not appear in the training data but needs to answer during testing. While reasoning models have a limited ability to learn these queries, it is implicit in the training process and leads to a severe impairment of performance. Therefore, it is beneficial to be proactive in discovering unseen FOL during training.
Secondly, the non-duality logic means the unreliable inverse mapping between entities. The inverse self-consistency of the logic is very rare in the noise-rich knowledge graph, and by learning this inverse mapping which we called non-duality logic, the model can achieve very strong robustness.

Recent studies (e.g., \cite{BetaE&KGreasoning}) have divided the seen / unseen logical paradigms to fairly evaluate models and \cite{noise,shao2021dskrl,wan2020adaptive,shan2018confidence} has used variational inference to handle potential noise in real-world knowledge graphs, none of them analyzed the implications of making these attempts on noisy knowledge graph in a holistic way. As a result, there is no holistic solution that can be specifically used to solve robust common sense reasoning problems on unseen logical paradigms. In this paper, we consider a plug-in module called \textit{Logic Diffusion (LoD)} to address these issues.

\textit{LoD} relies on a new structure called \textit{Hierarchical Conjunctive Query} which achieves a wide range of logic perception through different levels of unseen FOL discovery.
Then, with the help of \textit{Logic Specific Prompt}, \textit{LoD} can distinguish between different logical paradigms as well as learn the commonalities of the same kind of FOL.
However, due to the difference in the learning difficulty, the model tends to learn dominant or simple FOL, which is detrimental to the overall performance. Thus, we design a training mechanism called \textit{Gradient Adaption} in \textit{LoD} to slightly suppress the gradient on the fastest converging FOL and make the model have more opportunities to learn difficult or long-tailed samples in the early stage of training.
Figure~\ref{fig:1} is the illustration of the difference in FOL with or without \textit{LoD}.

%
Besides, \textit{LoD} is accompanied by a novel loss function to further achieve the robust logical diffusion when facing noisy data in training or testing sets.
The key idea of our loss function is extending the length of the mapping link and amplifying the perturbation, so that the noise data can be blocked. Following \cite{wan2020adaptive,han2020open}, we divide the noisy data of non-duality logic into (1) training with noise, and (2) testing with noise.
The model with our loss function tends to retain more FOL inputs that can be successfully reverse mapped in the training phase, and actively eliminates FOL inputs that cannot be successfully reverse mapped in the testing phase. 
Such a process ensures that the model has the ability to learn reverse mapping and remembers the useful data.
We use a mixed dataset to simulate what contains massive noise data of non-duality logic. In Section \ref{sec:exp}, we did detailed noise ratio experiments based on two commonly used public datasets FB15K \cite{TransE&FB15k&WN18} and NELL-995 \cite{NELL995}.

In summary, the main contributions of our work are three-fold:
\begin{itemize}
	\item We propose a plug-in module called \textit{Logic Diffusion (LoD)} which is extremely effective reasoning on both unseen and seen logical paradigms. To the best of our knowledge, \textit{LoD} is the first work to focus on logic diversity augmentation on logic perspective in knowledge graph reasoning by a flexible module.
    \item \textit{LoD} is accompanied by a novel loss function to further achieve the
robust logical diffusion when facing noisy data in training or testing
sets.
	\item Extensive experiments on four public datasets demonstrate the superiority of mainstream KG reasoning models with the proposed \textit{LoD} plug-in module over state-of-the-art. Moreover, our ablation study proves the general effectiveness of \textit{LoD} on the noise-rich knowledge graph.
\end{itemize}

%% file: sections/2_rela.tex
\section{Related Work}
Logical query reasoning on knowledge graphs has been recently received growing interest. Generally speaking, this work contains two main lines of works: how to model multi-hop relations and how to model numerous answers \cite{chen2020review,zhang2022knowledge}.

\subsection{Modeling Multi-hop Relations} 
Multi-hop logic reasoning try to answer queries with multi-hop logic permutations. Since embedding entities and relations in knowledge graph (KG) into low-dimensional vector space has been widely studied. Various works \cite{rossi2021knowledge,opencomp,nayyeri2021trans4e,emb1,emb2,emb3,emb4,emb5} can answer single-hop relational queries via link prediction but these models cannot handle complex logical reasoning. Therefore, to answer multi-hop FOL queries\cite{computegraph&multihop}, Graph Query Embedding (GQE) \cite{GQE} encodes conjunctive queries through a computation graph with relational projection and conjunction ($\wedge$) as operators. While path-based (i.e., deep reinforcement learning based) methods  \cite{NELL995,PRA,MINERVA,wan2021reasoning,dapath,wang2020adrl,li2021memorypath,chen2022rlpath} start from anchor entities and determine the answer set by traversing the intermediate entities via relational path and graph neural convolution based methods \cite{ye2022comprehensive,GNN1,GNN2,R-GCN,redGCNneib2,kosasih2022towards} pass message to iterate graph representation for reasoning. 


\subsection{Modeling Numerous Answers} 
Traditional knowledge graph reasoning works do not pay attention to potentially large sets of answer entities~\cite{Q2B}. That is, as long as one correct answer is inferred, KGR models is considered valid. However, it is unclear how such an entity set containing numerous answers can be represented as a single point in the vector space, causing inference inconsistent with the real situation.
So in order to handle numerous answer entities and inspired by metric learning, a series of works embeds queries into geometric shapes \cite{GQE,Q2B,ConE}, probability distributions \cite{PERM,line,proKGquery,probabilistic,vilnis2018probabilistic}, and complex objects \cite{quantum,faithful,particles}. Then by optimizing the similarity metrics between answer entities and queries, entities within border distance metrics of various representation spaces are regarded as correct answers.

However, above works lack of generalization to modeling queries of unseen logical paradigms and suffer from the interference of noisy data with non-duality logic, which are unavoidable in complex logical reasoning tasks on real-world knowledge graphs.


%% file: sections/3_pre.tex
\section{Preliminaries} 
\label{pre}
\subsection{Logic Format}
In the field of knowledge graph reasoning, most recent works focus on answering First-Order-logical queries rather than single-hop traversal within the triplet level. It is because answering FOL queries requires proper representation in both embedding and logic perspectives.
As it is illustrated in Figure~\ref{fig:1}, the query ``Who directs the film with the Academy Award for Best Original Score and the Golden Globe Award for Best Motion Picture?'' can be structured as a reasoning sub-graph $G_q$ \cite{computegraph&multihop}. Entities ``Golden Globe Award'' and ``Academy Award' are anchor entities, while entity ``Damien Chazelle'' is the target entity which refers to the answer, consisting a specific logical pattern of FOL. 

More specifically, First-Order Logic (FOL) is logic paradigms consisting of logical operators as conjunction ($\vee$), disconjunction ($\wedge$), universal quantification($\forall$), existential quantification ($\exists$) and negation ($\neg$) \footnote{Note that queries with universal quantification do not apply in real-world knowledge graphs since no entity connects with all the other entities. Furthermore, if it is necessary to introduce the universal quantifier into KG reasoning, the universal quantifier can be transformed from the existential quantification and the negation. Thus we will not discuss the queries with the universal quantifier.}.
Structured query-answer pairs of different FOL paradigms are the input of knowledge graph reasoning models during training while only queries during testing. Apparently it is impossible to exhaust all logical patterns. Moreover, in order to evaluate the generalization ability to unseen paradigms, paradigms for inference are more than training ones. While FOL paradigms without negation operators are Existential-Positive-First-order (EPFO), focused by some other works.
%
\subsection{Knowledge Graph Embedding}
Before logical reasoning, knowledge graph embedding (KGE) needs to map entities and relations in KG onto a representational latent space, which can participate in the former logical operations. Given a set of entities $\mathcal{E}$ and a set of relations $\mathcal{R}$, a knowledge graph $\mathcal{G} = \{ (e_s, r, e_o) \} \subset \mathcal{E} \times \mathcal{R} \times \mathcal{E}$ consists of factual triples as subject $e_s \in \mathcal{E}$, object $e_o \in \mathcal{E}$ and relational functions $r \in \mathcal{R}$ : $\mathcal{E} \times \mathcal{E} \rightarrow \{ \texttt{True, False} \}$ or other confidence and distributional metrics. 
Suppose ${\rm e}_s \in \mathbbm{R}^{d}, {\rm e}_o \in \mathbbm{R}^{d}, {\rm r} \in \mathbbm{R}^{d}$ are vector representations of subject $e_s$, object $e_o$ and relation $r$ in a triple of knowledge graphs, KGE works usually optimize their models according to a relation projection function ${\rm e}_o = f_r({\rm e}_s)$. As a result, the embedding features can be extracted from the embedding layer of pre-trained KGE models to computing the former logical operations. 

%
\subsection{Knowledge Graph Reasoning} 
As mentioned above, answering a FOL query $q$ can be simply illustrated as finding its answer set $[\![q]\!]$ according to its reasoning sub-graph $G_q$. A reasoning sub-graph is an abbreviation of the computation graph where nodes refer to entity sets and edges refer to logical operations. We call the starting point of FOL as anchor entities, and the end point of FOL as target entities.

Thus we can answer $q$ by executing logical operators from anchor entities. Based on this premise, logical operators can be matched according to the following rules:
\begin{itemize}
	\item{$ Relation \ Projection$} : 
	given a set $S \subseteq \mathcal{E} $ of entities and relation operator $r \in \mathcal{R} $, compute entities $\cup_{e \in S} P_r(e)$ 
	adjacent to $S$ via $r$: $P_r(e) \equiv \{ e^\prime \in \  : \ r(e, e^{\prime}) = \texttt{True} \}$.
	\item{$Intersection$} : Given $n$ sets $\{ S_1, S_2, \ldots, S_n\}$ of entities, compute their intersection $\cap_{i = 1}^n S_i$.
	\item{$Union$} : Given $n$ sets $\{ S_1, S_2, \ldots, S_n\}$ of entities, compute their union $\cup_{i = 1}^n S_i$.
	\item{$Negation$} : Given a set $S \subseteq \ $ of entities , compute its complement $\overline{S} \equiv \mathcal{E} \:\backslash\: S$.
\end{itemize}

In a word, knowledge graph reasoning mainly aims at answering FOL queries by executing several logical operators with vector representations after embedding. 

%% file: sections/4_method.tex
\begin{figure}
	\begin{center}
        \includegraphics[width=1\linewidth]{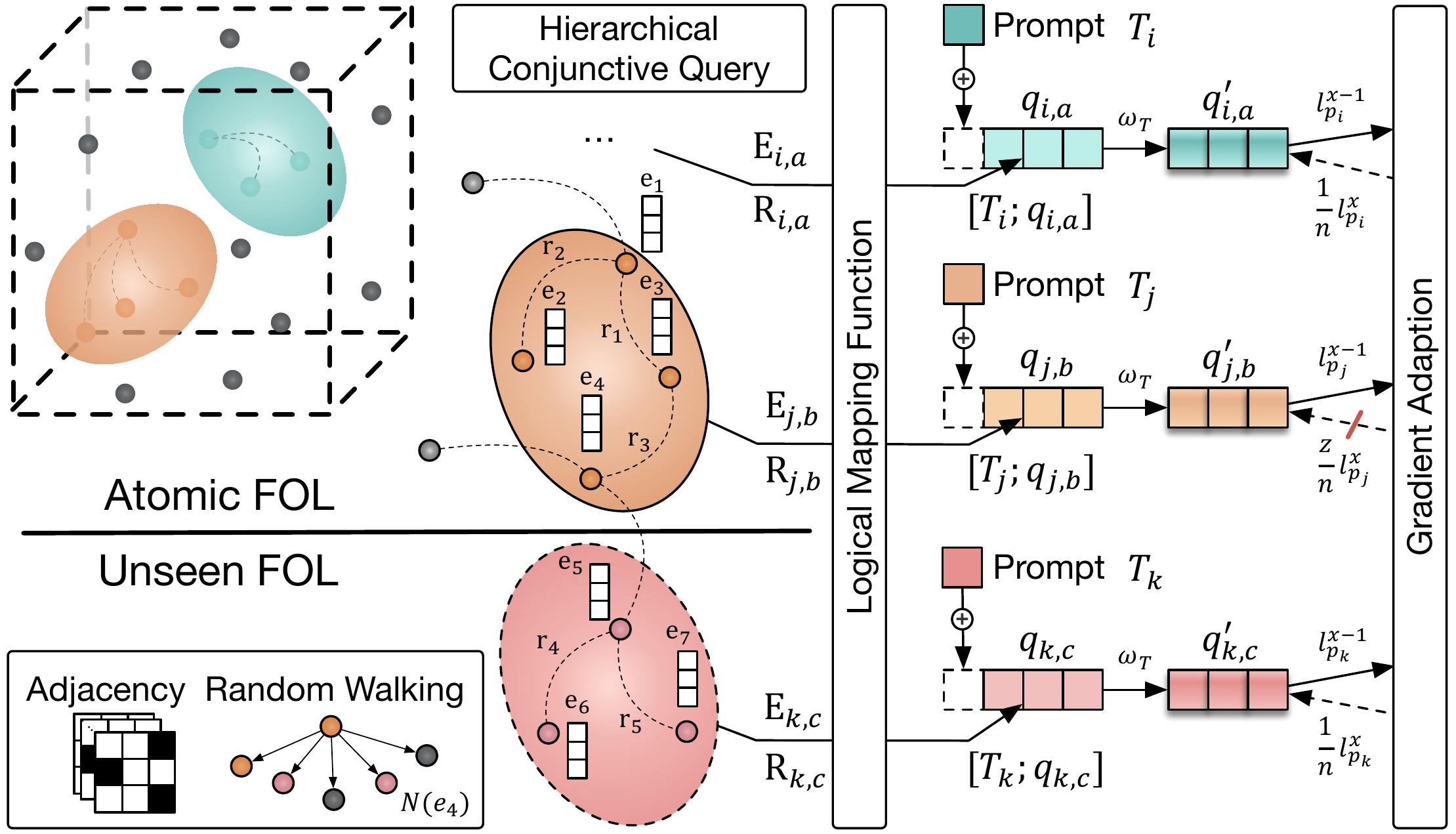}
	\end{center}
	\caption{Overall framework Logic Diffusion architecture. Along the direction of data flow, there are Hierarchical Conjunctive Query, Logic Specific Prompt and Gradient Adaption, discovering and learning multiple FOL paradigms.}
    \label{fig:2}
\end{figure}
\begin{figure}
	\begin{center}
 	\includegraphics[width=0.9\linewidth]{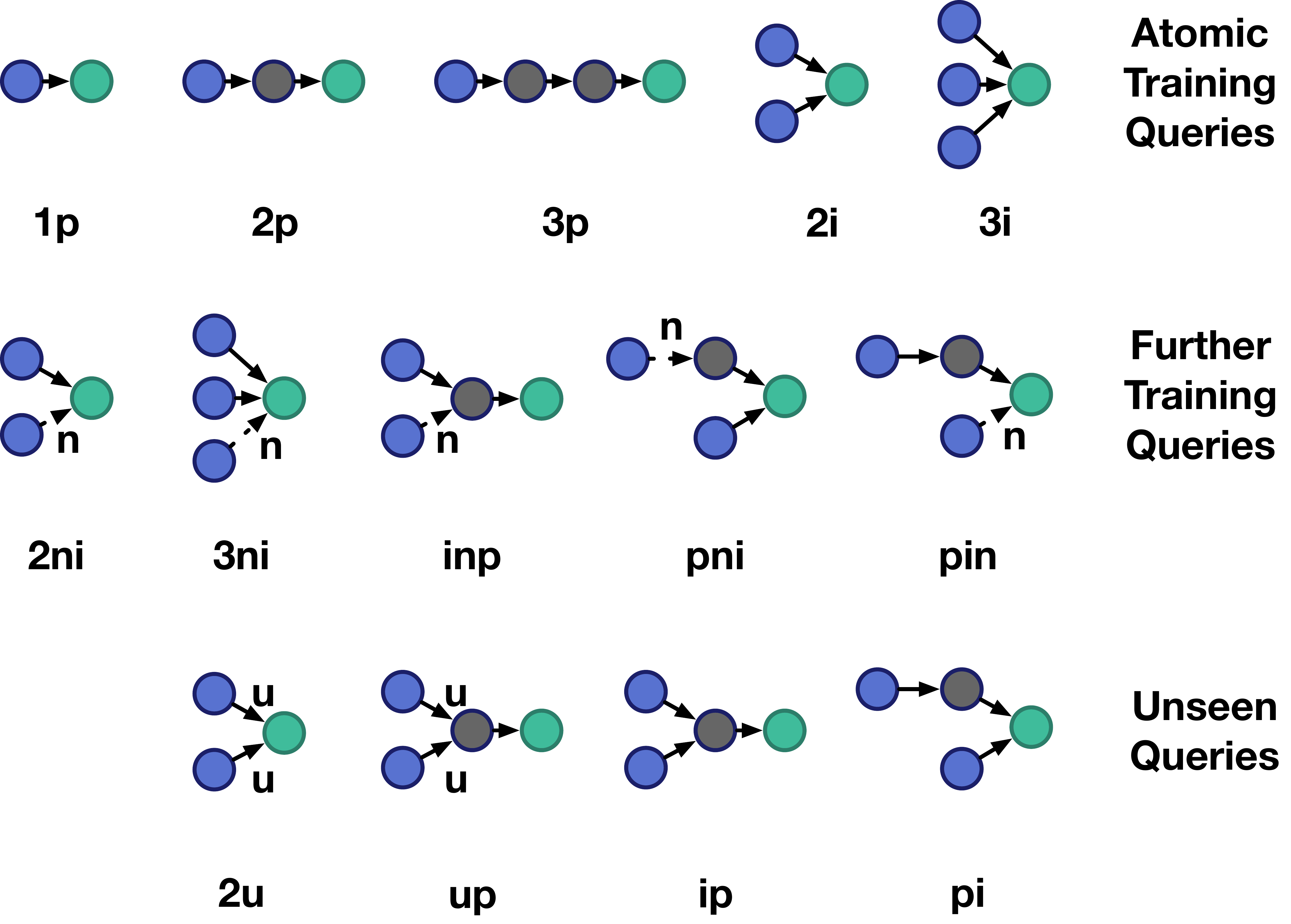}
	\end{center}
	\caption{Illustration of 14 typical FOL paradigms, including atomic training, further training and unseen queries.}
	\label{fig:3}
\end{figure}
\begin{figure}
	\begin{center}
 	\includegraphics[width=0.8\linewidth]{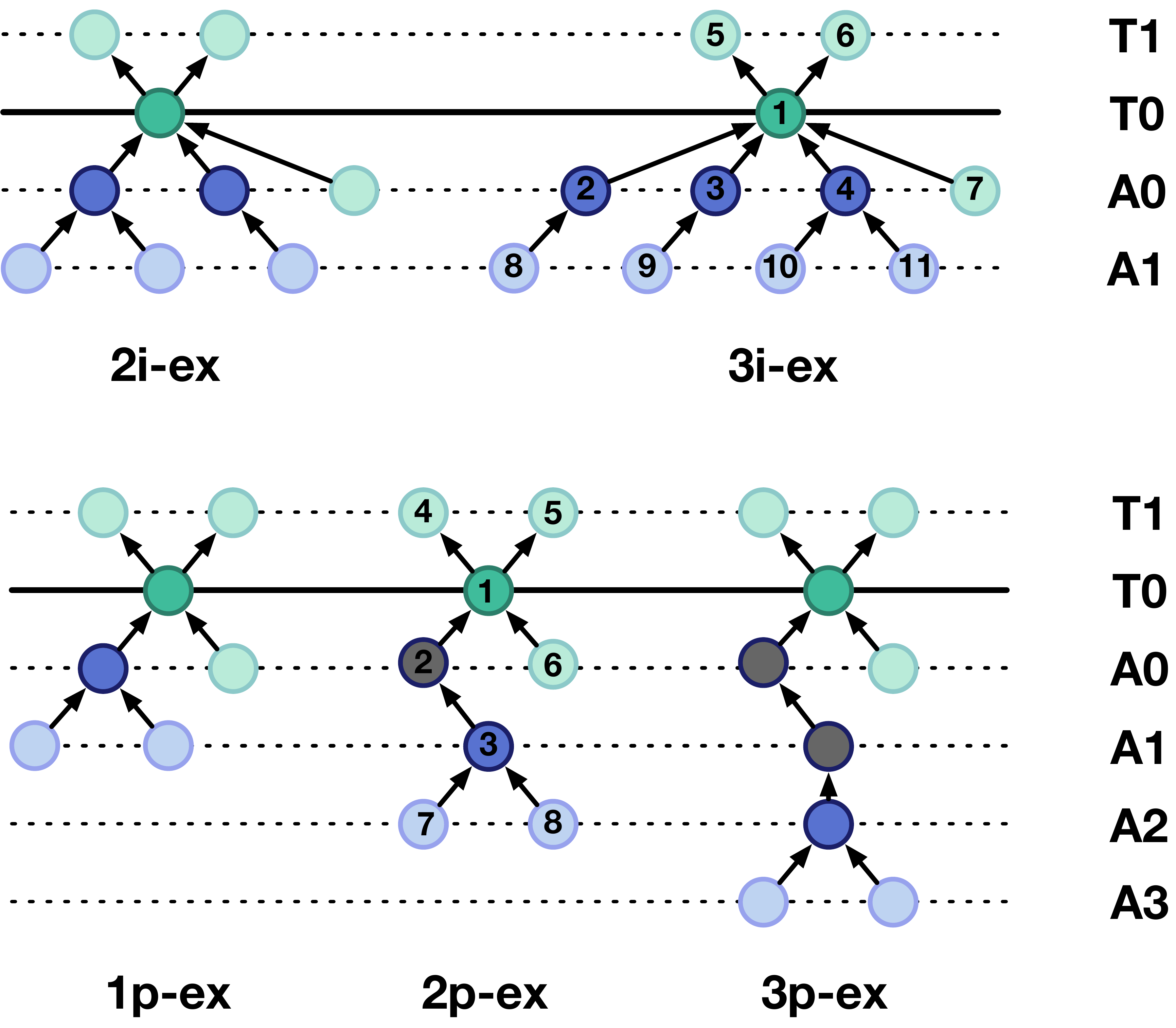}
	\end{center}
	\caption{Illustration of the Hierarchical Conjunctive Query. Each group represents a FOL paradigm. Nodes ${\rm T0}$ layer contains the positive target entity $t$ which indicates the answer of a certain FOL. $e_i$ is one layer below $e_j$ if there exists $(e_i, e_j)$ satisfying $r(e_i, e_j) = \texttt{True}$ where $r$ is the relation mapping.}
	\label{fig:4}
\end{figure}
\section{Logic Diffusion}
In this section, we introduce the proposed plug-in module——\textit{Logic Diffusion} in detail. \textit{LoD} accepts any permutation and combination of FOL instances of clean or noise-rich KG as input. Referring to Figure~\ref{fig:2} and Figure~\ref{fig:3}, along the direction of data flow, \textit{LoD} has two main designs including \textit{LoD} architecture and adaptive loss. \textit{LoD} architecture, consisting of $[$Hierarchical Conjunctive Query$]$, $[$Logic Specific Prompt$]$ and $[$Gradient Adaption$]$ is used to discover and learn multiple FOL paradigms. Note that the logic diffusion process mainly happens during $[$Hierarchical Conjunctive Query$]$, which extends logical diversity by random walking among the entity distribution; $[$Logic Specific Prompt$]$ specializes the feature within each FOL paradigms while distinguishes among different ones; $[$Gradient Adaption$]$ dynamically adjusts convergence speed and achieves adaptively balance of several logic patterns. While \textit{LoD Loss} is used to keep duality logic during robust reasoning and release the interference of noisy data in real-world KG. 

\subsection{LoD Architecture}
\noindent\textbf{Hierarchical Conjunctive Query.}
In this section, we introduce Hierarchical Conjunctive Query, which is the core component of logical diffusion, discovering unseen logic paradigms and sample sub-logic patterns both in abstract reasoning sub-graphs and figurative instances.
To implement Logic Diffusion, we firstly define Hierarchical Conjunctive Query as a theoretical guide. Following \cite{BetaE&KGreasoning}, we selected 14 typical FOL paradigms illustrated in Figure~\ref{fig:3} where atomic training queries and further training queries denote the input samples of KGR models, while unseen queries denote samples do not participate training process but evaluation. Though there are far more FOL paradigms, we can still learn the generalization ability of KGR models through modeling these unseen queries in this specific setting. As the name suggests, Hierarchical Conjunctive Query has a hierarchical form, from the bottom to top which is:

\begin{itemize}
	\item{$Bottom \ Layer$} : Atomic Query
 \end{itemize}

Following \cite{BetaE&KGreasoning}, there are 14 typical FOL paradigms including atomic training $\{1p, 2p, 3p, 2i, 3i\}$, further training $\{2ni, 3ni, inp,$ $ pni, pin\}$ and unseen $\{ip, pi, 2u, up\}$. Since the atomic training queries $\{1p, 2p, 3p, 2i, 3i\}$ are raw and simple input samples, we simply take the them as the bottom layer of Hierarchical Conjunctive Query.

\begin{itemize}
	\item{$Middle \ Layer$} : Sub-Graph Diffusion
 \end{itemize}

The middle layer of Hierarchical Conjunctive Query contains the extensions and permutations of Atomic Queries, which is theoretically infinite (e.g., $\{9p, 10i, 7i3p\dots\}$). As shown in Figure~\ref{fig:4}, given a query $q$, reasoning sub-graph $G_q$ and positive target $t$ of an atomic query, we diffuse $G_q$ by neighbors. Suppose $N(e)\equiv\{e_{neib}|r(e_{neib},e)\vee r(e,e_{neib}),r\in\mathcal{R}\}$ is the collection of neighbors of each central node $e$, we get $N(e_1), N(e_2), \dots, N(e_n)$ and $N(t)$. If $r({e_{neib}}^j, e_i) = \texttt{True}$ or ${t_{neib}}^j \in N(t)$, we associate ${e_{neib}}^j$ and ${t_{neib}}^j$ to $G_q$ via $r$. We stipulate that ${e_{neib}}_j$ is different from any other entities in $q$ and $t$.

\begin{itemize}
	\item{$Top \ Layer$} : Unseen Query
 \end{itemize}

The top layer contains unseen FOL which are not present in the training set. It is generated by random walking in the diffused reasoning sub-graphs to cover complex logic in the real-world knowledge graph, i.e., they are not given directly with the training set. 
Here we give examples of $\rm 3i-ex$ and $\rm 2p-ex$. From $\rm 3i-ex$, we can retrieve a sub-structure (nodes 1, 2, 3, 5 and relational edges attached, etc.) corresponding to ${\rm ic}$. From $\rm 2p-ex$ , we can retrieve a sub-structure (nodes 1, 2, 3, 6 and relational edges attached, etc.) corresponding to ${\rm ci}$. Note that each node in the diagram actually represents a collection of entities, which is only represented as a single entity to simplified.

In the paradigm $\rm 3i-ex$, for ${\rm ic}$ query with nodes 1, 2, 3, 5, we label nodes 2, 3 as the new anchor entities, and node 5 as the new positive target entity. Then we randomly sample $n\_s$ entities $\notin N(1)$ as the negative target entities where $N(1)$ denotes the collection of neighbors of node 1.

In paradigm $\rm 2p-ex$, for ${\rm ci}$ query with nodes 1, 2, 3, 6, note since node 2 is unknown to models, we will not choose this kind of nodes as the new target nor anchor entities. Similarly we label nodes 2, 6 as the new anchor entities, node 5 as the new positive target entities, and ones $\notin N(6)$ as the negative target entities.

\noindent\textbf{Logic Specific Prompt.}
Stemming from recent advances in natural language processing, prompt learning initially fills the input sample into properly handcrafted prompt templates, so that a pre-trained language model can ``understand'' the task~\cite{Brown:LM-fS-learners}. Similarly, we define different kinds of FOL as different tasks and propose Logic Specific Prompt to make the learning of the model more targeted.
Specifically, given a set $\{Q_1, Q_2, \ldots, Q_k\}$ belonging to $k$ kinds of FOL, we generate $k$ random vectors $\{T_1, T_2, \ldots, T_k\}$ from normal distribution as initialized prompt. Then we concatenate prompt $T_i$ with corresponding ${\rm q}_{i,j}$ as ${\rm q}_{i,j}^\prime$. The formula of ${\rm q}_{i,j}^\prime$ is as follow:
\begin{equation}
    {\rm q}_{i,j}^\prime = {\rm ReLU}(\omega_{\rm T} \left [{\rm T_i};{\rm q}_{i,j}\right ]+{\rm b}_{\rm T}) \label{}
\end{equation}
where $T_i \in \mathbbm{R}^{d_{\rm T}} $ in which $d_T$ denotes the dimension of prompt, filter $\omega_{\rm T} \in \mathbbm{R}^{(d+d_{\rm T})\times d} $, ${\rm b}_{\rm T} \in \mathbbm{R}^d$ is the bias. The optimal value of $d_T$ is 32.

\noindent\textbf{Gradient Adaption.}
Our approach involves a rich variety of FOL due to the introduction of Hierarchical Conjunctive Query. If no restrictions are imposed, the model will tend to learn dominant or simple FOL, which is detrimental to the overall performance. The key idea of Gradient Adaption is to slightly suppress the gradient on the fastest converging FOL. Through this processing, the model has more opportunities to learn difficult or long-tailed samples in the early stage of training. Specifically, suppose $l_p^{i-1} = \{l_{p_1}^{i-1}, l_{p_2}^{i-1}, \ldots, l_{p_n}^{i-1}\}$ is the calculated loss value of $\{p_1, p_2,\ldots, p_n\}$ at the $(i\texttt{-}1)$-th iteration, $l_p^{i} = \{l_{p_1}^{i}, l_{p_2}^{i}, \ldots, l_{p_n}^{i}\}$ at the $i$-th iteration. The suppression of $l_{p}^i$ is computed as follow:
\begin{equation}
    {l_{p}^i}^\prime =\frac{1}{n}(\sum^n\limits_{j=1}l_{p_j}^i - z\cdot l_{p_{\zeta}}^i)
\end{equation}
The attenuation factor $z$ is 0.05. $\zeta = \mathop{\mathrm{argmax}}\limits_{n\in \mathcal{R}}{l_{p_n}^{i-1}}$.

\begin{figure}
	\begin{center}
 	\includegraphics[width=0.65\linewidth]{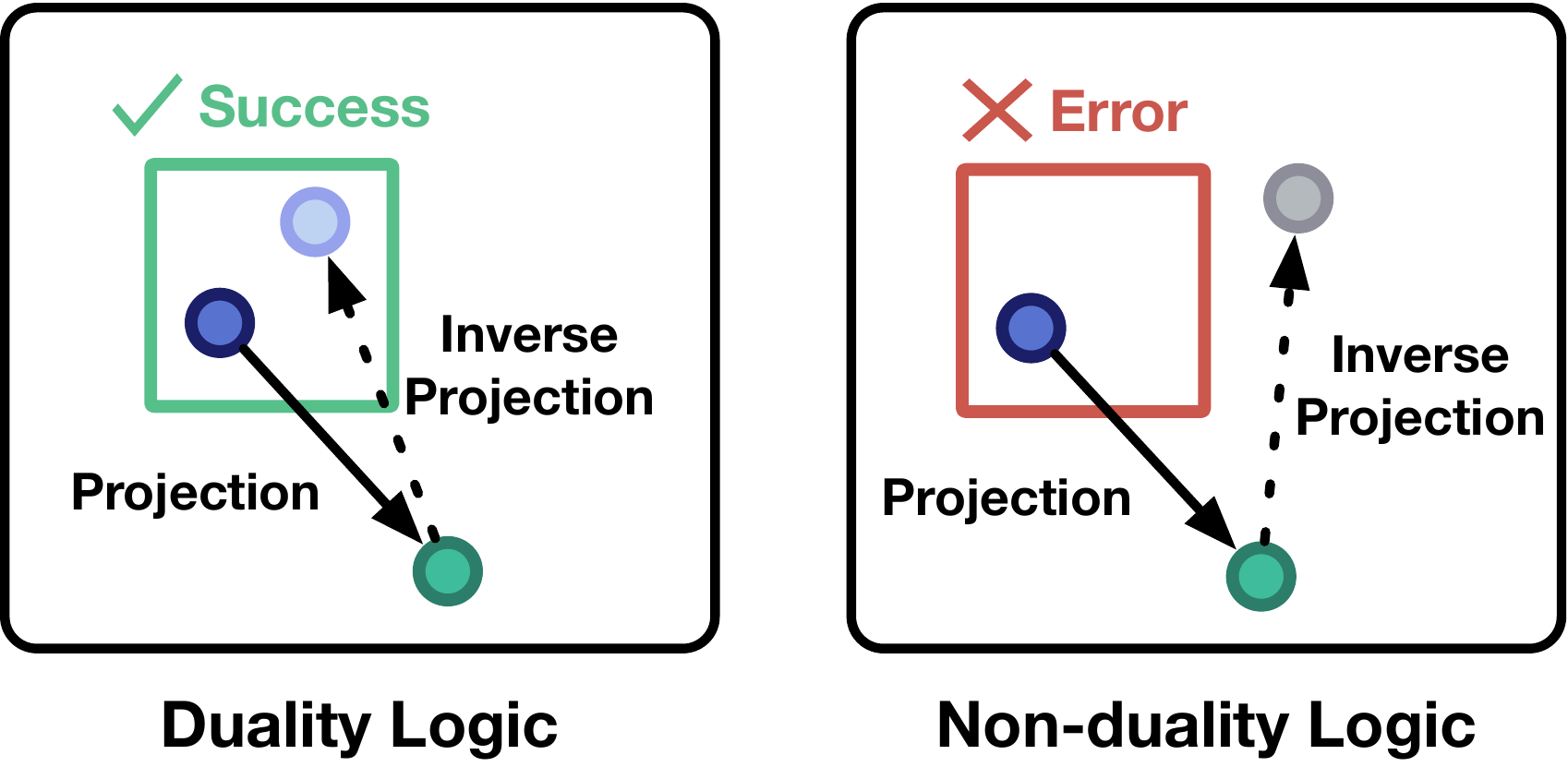}
	\end{center}
	\caption{Logic Diffusion Loss can perfect duality logic (i.e., bidirectional mapping) between queries and answers in noise-rich KG. Thus non-duality noisy data can be distinguished by a distance threshold.}
	\label{fig:5}
\end{figure}

\subsection{LoD Loss}
\label{sec:medloss}
In retrospect, we disassemble the noise-rich knowledge graph into two major problems: unseen logic paradigms and noisy data with non-duality logic. In this section, we will introduce our solution for non-duality logic. 
As introduced in Section~\ref{pre}, a complex query constructed with permutation of logic operators can be simply illustrated as a reasoning sub-graph $G_q$. Answer queries by executing logic operators from anchor entities to target entities is defined as a forward reasoning procedure. 
However, the existing methods use only forward procedure which leads to unreliable bidirectional mapping between queries and answers, which is non-duality. Indeed, how to design a backward reasoning procedure and perfect the inference-inversion process are the keys to solve this problem. That is, to keep duality logic in knowledge graph reasoning.
Specifically, given an anchor entity set $\{a_1, a_2, \ldots, a_s\}$ of an FOL query $q$, we define positive / negative forward relational paths $\varrho $ / $\varrho_i^\prime$ in the reasoning sub-graph $G_q$. $\varrho $ starts from a certain anchor entity $a_j$ to a positive target entity $t \in [\![q]\!] $ through relation mappings. $\varrho_i^\prime$ starts from $a_j$ to a random negative entity $t_i^\prime \notin [\![q]\!]$. Apparently, the only kind of logic operators along a positive / negative forward relational path is the relation projection $P_r(e) \equiv \{e^{\prime} \in \ : \ r(e, e^{\prime}) = \texttt{True}\}$. After defining forward relational paths, we generate backward relational paths with a similar strategy. We replace all relation projection operators in a forward relational path with their inverse mapping $P_r^{-1}(e) \equiv \{e^{\prime} \in \ : \ r(e^{\prime}, e) = \texttt{True}\}$. $\varrho ^{-1}_j$ denotes a positive backward relational path from $t$ to $a_j$. ${\varrho^{-1}}_{i,j}^\prime$ denotes a negative backward relational path from $t_i^\prime$ to $a_j$. $\mathcal{L}_{dl}^{j}$ of the anchor entity $a_j$ is:
\begin{equation}
    \begin{aligned}
    \mathcal{L}_{dl}^{j} =  & -{\rm log} \sigma ({\rm score}({\rm a}_j, {{\rm \varrho }^{-1}}_j)) \\
     & - \sum_{i=1}^{n\_s} {\rm log} \sigma (-{\rm score}({\rm a}_j, {{\rm \varrho}^{-1}}_{i,j}^\prime))
    \end{aligned}
\label{}
\end{equation}
Accordingly, we define \textit{LoD Loss} as follows: 
\begin{equation}
    \mathcal{L}_{dl} =\left\{
    \begin{aligned}
    \ {\rm Max}(\mathcal{L}_{dl}^1, \mathcal{L}_{dl}^2, \mathcal{L}_{dl}^3) &, \ \  {\rm if} \ p \in \{{\rm 3i\}}    \\    
    \ {\rm Max}(\mathcal{L}_{dl}^1, \mathcal{L}_{dl}^2) &, \ \  {\rm if} \ p \in \{{\rm 2i,\ ip,\ pi\}}    \\
    \ {\rm Min}(\mathcal{L}_{dl}^1, \mathcal{L}_{dl}^2) &, \ \  {\rm if} \ p \in \{{\rm 2u,\ up\}}    \\
    \ \mathcal{L}_{dl}^1 &, \ \  {\rm if} \ p \in \{{\rm 1p, 2p, \ 3p\}}
    \end{aligned}
    \right.
\end{equation}
The noisy data with non-duality logic can be distinguished by ${\rm score}({\rm \varrho }^{-1}, {\rm a})$ with the distance threshold $\varsigma$.

In addition to $\mathcal{L}_{dl}$, we also need to use a contrast loss $\mathcal{L}_r$ to optimize the model by pulling $(\rm q,\rm t)$ closer while pushing $(\rm q,{\rm t}_i^\prime)$ farther ($q$ is query, $t \in [\![q]\!]$ is target and ${\rm t}_i^\prime \notin [\![q]\!]$ is random negative target). The formula is as follow:
\begin{equation}
\begin{aligned}
    \mathcal{L}_r = & -{\rm log} \sigma ({\rm score}({\rm t}, {\rm q }))\\
    &- \sum_{i=1}^{n\_s} {\rm log} \sigma (-{\rm score}({\rm t}_i^\prime, {\rm q })) 
\end{aligned}
\label{}
\end{equation}
where ${\rm score}$ denotes a normalized score function to evaluate the similarity between an entity and a query. $\sigma$ is the sigmoid function, and $n\_s$ denotes the negative sample size. 
Therefore, the overall loss function $\mathcal{L}$ can be calculated as follow:
\begin{equation}
    \mathcal{L} = \mathcal{L}_r + \lambda \mathcal{L}_{dl}
    \label{}
\end{equation}
where $\lambda \in [0, 1] $ is a hyperparameter. The best is 0.8.

%% file: sections/5_exp.tex
\begin{table}[t]
\centering
\setlength\tabcolsep{2pt}
\setlength{\abovecaptionskip}{2pt}
\caption{Statistics of datasets as well as training, validation and test edge splits. \textit{Entity} means the number of entities. \textit{Rela.} means the number of relations. \textit{Tr-Edge} means the number of edges in training set. \textit{V-Edge} means the number of edges in validation set. \textit{Ts-Edge} means the number of edges in testing set. \textit{Ts-Edge} means the number all edges.}
\scalebox{0.95}{
\begin{tabular}{@{}lcccccc@{}}
\toprule
Dataset   & Entity & Rela. & Tr-Edge & V-Edge & Ts-Edge & Edge \\ \midrule
FB15k     & 14,951   & 1,345     & 483,142        & 50,000           & 59,071     & 592,213     \\
FB15k-237 & 14,505   & 237       & 272,115        & 17,526           & 20,438     & 310,079     \\
NELL995   & 63,361   & 200       & 114,213        & 14,324           & 14,267     & 142,804     \\
WN18      & 40,493   & 18        & 141,442        & 5,000            & 5,000      & 151,442     \\ \bottomrule
\end{tabular}}
\label{table:a}
\end{table}

\begin{table}[h]
\centering
\setlength{\abovecaptionskip}{2pt}
\caption{The best hyperparameters of \textit{LoD}.}
\begin{tabular}{@{}ccccccc@{}}
\toprule
$d$ & ${\rm lr}$ & $b_z$ & $n\_s$ & $M$ & $d_T$ & $z$ \\ \midrule
768 & 0.0005     & 512        & 128    & 64  & 32    & 0.05 \\ \bottomrule
\end{tabular}
\label{table:ap2}
\end{table}

\begin{table*}[t]
\renewcommand\arraystretch{0.8}
\centering
\setlength\tabcolsep{4pt}
\setlength{\abovecaptionskip}{2pt}%
\caption{MRR results (\%) on answering FOL queries of raw methods and their enhanced versions (with \textit{LoD}) over FB15k. ${\rm avg}$ denotes the the average MRR on all queries (i.e., FOL), ${\rm avg}_p$ on EPFO, and ${\rm avg}_{unseen}$ on unseen paradigms. 
Note that limited by training time, we manually reproduce the Logical Mapping Function of FuzzQE as it was described in the paper and marked as FuzzQE*.}
\scalebox{0.95}{
\begin{tabular}{@{}lcccccccccccccccccc@{}}
\toprule[1pt]
Model                       &\textit{LoD} & ${\rm avg}$ & ${\rm avg}_p$ & ${\rm avg}_{unseen}$ & ${\rm 1p}$ & ${\rm 2p}$ & ${\rm 3p}$ & ${\rm 2i}$ & ${\rm 3i}$ & ${\rm pi}$ & ${\rm ip}$ & ${\rm 2u}$ & ${\rm up}$ & ${\rm 2in}$ & ${\rm 3in}$ & ${\rm inp}$ & ${\rm pin}$ & ${\rm pni}$ \\ \midrule 
\\ [-2ex]\multicolumn{19}{c}{FB15k} 
\\ [-0.1ex]\midrule \\[-2.4ex]
                            & \small{\base{\XSolidBrush}}                                    & -                                   & 28.0             &20.0                     & 55.3                               & 15.2                               & 11.2                               & 39.1                               & 51.0                               & 27.7                               & 18.8                               & 22.1                               & 11.5                               & -                                   & -                                   & -                                   & -                                   & -                                                    \\
\multirow{-2}{*}{GQE}       & \small{\checkc{\Checkmark}}                                 & -           & \textbf{30.6}   & \textbf{22.8}       & \cellcolor[HTML]{EFEFEF}57.9       & \cellcolor[HTML]{EFEFEF}16.6       & \cellcolor[HTML]{EFEFEF}13.7       & \cellcolor[HTML]{EFEFEF}42.3       & \cellcolor[HTML]{EFEFEF}54.1       & \cellcolor[HTML]{EFEFEF}31.6       & \cellcolor[HTML]{EFEFEF}21.3       & \cellcolor[HTML]{EFEFEF}24.8       & \cellcolor[HTML]{EFEFEF}13.4       & \cellcolor[HTML]{EFEFEF}-           & \cellcolor[HTML]{EFEFEF}-           & \cellcolor[HTML]{EFEFEF}-           & \cellcolor[HTML]{EFEFEF}-           & \cellcolor[HTML]{EFEFEF}-                            \\ 
                            & \small{\base{\XSolidBrush}}                                    & -                                   & 37.9       & 29.2                          & 67.2                               & 21.7                               & 14.3                               & 54.9                               & 66.3                               & 39.7                               & 25.9                               & 34.9                               & 16.5                               & -                                   & -                                   & -                                   & -                                   & -                                                    \\
\multirow{-2}{*}{Query2Box} & \small{\checkc{\Checkmark}}                                 & -           & \textbf{40.4}     & \textbf{31.7}     & \cellcolor[HTML]{EFEFEF}70.9       & \cellcolor[HTML]{EFEFEF}24.0       & \cellcolor[HTML]{EFEFEF}16.3       & \cellcolor[HTML]{EFEFEF}57.1       & \cellcolor[HTML]{EFEFEF}68.5       & \cellcolor[HTML]{EFEFEF}42.3       & \cellcolor[HTML]{EFEFEF}28.4       & \cellcolor[HTML]{EFEFEF}37.5       & \cellcolor[HTML]{EFEFEF}18.4       & \cellcolor[HTML]{EFEFEF}-           & \cellcolor[HTML]{EFEFEF}-           & \cellcolor[HTML]{EFEFEF}-           & \cellcolor[HTML]{EFEFEF}-           & \cellcolor[HTML]{EFEFEF}-                            \\
                            & \small{\base{\XSolidBrush}}                                    & -                                   & 48.1           & 40.1                       & 73.1                               & 28.2                               & 26.3                               & 65.4                               & 79.6                               & 48.3                               & 40.1                               & 43.9                               & 28.1                               & -                                   & -                                   & -                                   & -                                   & -                                                    \\
\multirow{-2}{*}{ABIN}      & \small{\checkc{\Checkmark}}                                 & -           & \textbf{49.3}    & \textbf{41.2}      & \cellcolor[HTML]{EFEFEF}74.8       & \cellcolor[HTML]{EFEFEF}29.1       & \cellcolor[HTML]{EFEFEF}26.9       & \cellcolor[HTML]{EFEFEF}66.5       & \cellcolor[HTML]{EFEFEF}81.2       & \cellcolor[HTML]{EFEFEF}48.7       & \cellcolor[HTML]{EFEFEF}42.5       & \cellcolor[HTML]{EFEFEF}45.0       & \cellcolor[HTML]{EFEFEF}28.7       & \cellcolor[HTML]{EFEFEF}-           & \cellcolor[HTML]{EFEFEF}-           & \cellcolor[HTML]{EFEFEF}-           & \cellcolor[HTML]{EFEFEF}-           & \cellcolor[HTML]{EFEFEF}-                            \\
                            & \small{\base{\XSolidBrush}}                                    & 31.1                                & 41.7          &34.3                        & 65.7                               & 26.0                               & 24.9                               & 55.4                               & 66.3                               & 44.3                               & 27.8                               & 39.8                               & 25.2                               & 14.1                                & 14.6                                & 11.5                                & 6.6                                 & 12.6                                                 \\
\multirow{-2}{*}{BetaE}     & \small{\checkc{\Checkmark}}                                 & \textbf{32.6}        & \textbf{43.7}      &\textbf{37.0}    & \cellcolor[HTML]{EFEFEF}67.4       & \cellcolor[HTML]{EFEFEF}27.0       & \cellcolor[HTML]{EFEFEF}25.7       & \cellcolor[HTML]{EFEFEF}56.9       & \cellcolor[HTML]{EFEFEF}68.0       & \cellcolor[HTML]{EFEFEF}47.6       & \cellcolor[HTML]{EFEFEF}30.3       & \cellcolor[HTML]{EFEFEF}41.7       & \cellcolor[HTML]{EFEFEF}28.4       & \cellcolor[HTML]{EFEFEF}15.4        & \cellcolor[HTML]{EFEFEF}15.1        & \cellcolor[HTML]{EFEFEF}12.2        & \cellcolor[HTML]{EFEFEF}7.1         & \cellcolor[HTML]{EFEFEF}13.7                         \\
                            & \small{\base{\XSolidBrush}}                                   & 31.5                                & 41.9             &33.0                     & 67.9                               & 27.2                               & 25.9                               & 56.2                               & 67.7                               & 46.7                               & 30.2                               & 34.6                               & 20.6                               & 15.2                                &15.7                                     & 12.4                                & 7.3                                 & 13.5                                                 \\
\multirow{-2}{*}{FuzzQE*}   & \small{\checkc{\Checkmark}}                                 & \textbf{33.1}                                & \textbf{43.8} & \textbf{35.8}          & \cellcolor[HTML]{EFEFEF}69.2       & \cellcolor[HTML]{EFEFEF}28.2       & \cellcolor[HTML]{EFEFEF}26.3       & \cellcolor[HTML]{EFEFEF}57.9       & \cellcolor[HTML]{EFEFEF}69.4       & \cellcolor[HTML]{EFEFEF}48.0       & \cellcolor[HTML]{EFEFEF}31.5       & \cellcolor[HTML]{EFEFEF}38.5       & \cellcolor[HTML]{EFEFEF}25.2       & \cellcolor[HTML]{EFEFEF}16.4        & \cellcolor[HTML]{EFEFEF}17.1        & \cellcolor[HTML]{EFEFEF}13.5        & \cellcolor[HTML]{EFEFEF}8.0         & \cellcolor[HTML]{EFEFEF}{14.3}                         \\ \midrule
\\ [-2ex]\multicolumn{19}{c}{FB15k-237} 
\\ [-0.1ex]\midrule \\[-2.4ex]
                            & \small{\base{\XSolidBrush}} & -             & 16.4     & 10.3     & 35.6                         & 7.4                          & 5.4                          & 23.3                         & 34.5                         & 16.6                         & 10.5                         & 8.3                          & 5.8                          & -                           & -                            & -                           & -                           & -                           \\
\multirow{-2}{*}{GQE}       & \small{\checkc{\Checkmark}}   & -      &\textbf{12.3}       & \textbf{18.4} & \cellcolor[HTML]{EFEFEF}38.1 & \cellcolor[HTML]{EFEFEF}9.0  & \cellcolor[HTML]{EFEFEF}6.8  & \cellcolor[HTML]{EFEFEF}25.4 & \cellcolor[HTML]{EFEFEF}36.9 & \cellcolor[HTML]{EFEFEF}19.3 & \cellcolor[HTML]{EFEFEF}12.7 & \cellcolor[HTML]{EFEFEF}10.1 & \cellcolor[HTML]{EFEFEF}7.1  & \cellcolor[HTML]{EFEFEF}-   & \cellcolor[HTML]{EFEFEF}-    & \cellcolor[HTML]{EFEFEF}-   & \cellcolor[HTML]{EFEFEF}-   & \cellcolor[HTML]{EFEFEF}-   \\
                            & \small{\base{\XSolidBrush}} & -             & 20.6     & 14.0     & 41.0                         & 9.7                          & 7.0                          & 29.4                         & 42.4                         & 21.3                         & 12.6                         & 12.4                         & 9.6                          & -                           & -                            & -                           & -                           & -                           \\
\multirow{-2}{*}{Query2Box} & \small{\checkc{\Checkmark}}   & -             & \textbf{22.9} &\textbf{16.3} & \cellcolor[HTML]{EFEFEF}44.2 & \cellcolor[HTML]{EFEFEF}11.4 & \cellcolor[HTML]{EFEFEF}8.7  & \cellcolor[HTML]{EFEFEF}32.2 & \cellcolor[HTML]{EFEFEF}44.5 & \cellcolor[HTML]{EFEFEF}24.1 & \cellcolor[HTML]{EFEFEF}15.3 & \cellcolor[HTML]{EFEFEF}14.8 & \cellcolor[HTML]{EFEFEF}11.0 & \cellcolor[HTML]{EFEFEF}-   & \cellcolor[HTML]{EFEFEF}-    & \cellcolor[HTML]{EFEFEF}-   & \cellcolor[HTML]{EFEFEF}-   & \cellcolor[HTML]{EFEFEF}-   \\
                            & \small{\base{\XSolidBrush}} & -             & 30.2     &21.7     & 54.9                         & 17.2                         & 14.1                         & 42.4                         & 56.2                         & 31.4                         & 18.2                         & 19.9                         & 17.1                         & -                           & -                            & -                           & -                           & -                           \\
\multirow{-2}{*}{ABIN}      & \small{\checkc{\Checkmark}}   & -             & \textbf{32.1} & \textbf{23.2} & \cellcolor[HTML]{EFEFEF}58.5 & \cellcolor[HTML]{EFEFEF}18.3 & \cellcolor[HTML]{EFEFEF}15.0 & \cellcolor[HTML]{EFEFEF}44.6 & \cellcolor[HTML]{EFEFEF}59.8 & \cellcolor[HTML]{EFEFEF}33.6 & \cellcolor[HTML]{EFEFEF}19.9 & \cellcolor[HTML]{EFEFEF}21.2 & \cellcolor[HTML]{EFEFEF}18.0 & \cellcolor[HTML]{EFEFEF}-   & \cellcolor[HTML]{EFEFEF}-    & \cellcolor[HTML]{EFEFEF}-   & \cellcolor[HTML]{EFEFEF}-   & \cellcolor[HTML]{EFEFEF}-   \\
                            & \small{\base{\XSolidBrush}} & 15.5          & 21      & 14.3      & 39.3                         & 11.0                         & 10.0                         & 28.8                         & 42.6                         & 22.4                         & 12.7                         & 12.5                         & 9.7                          & 5.1                         & 8.0                          & 7.4                         & 3.5                         & 3.4                         \\
\multirow{-2}{*}{BetaE}     & \small{\checkc{\Checkmark}}   & \textbf{17.1} & \textbf{23.1} &\textbf{16.2} & \cellcolor[HTML]{EFEFEF}42.5 & \cellcolor[HTML]{EFEFEF}12.3 & \cellcolor[HTML]{EFEFEF}11.1 & \cellcolor[HTML]{EFEFEF}31.0 & \cellcolor[HTML]{EFEFEF}45.7 & \cellcolor[HTML]{EFEFEF}25.7 & \cellcolor[HTML]{EFEFEF}13.4 & \cellcolor[HTML]{EFEFEF}14.6 & \cellcolor[HTML]{EFEFEF}11.2 & \cellcolor[HTML]{EFEFEF}6.7 & \cellcolor[HTML]{EFEFEF}8.7  & \cellcolor[HTML]{EFEFEF}8.7 & \cellcolor[HTML]{EFEFEF}4.2 & \cellcolor[HTML]{EFEFEF}4.1 \\
                            & \small{\base{\XSolidBrush}} & 17.1          & 22.8      & 16.6    & 40.7                         & 11.8                         & 9.8                          & 31.4                         & 45.5                         & 25.0                         & 17.4                         & 13.9                         & 9.9                          & 7.3                         & 10.2                         & 6.8                         & 4.9                         & 5.2                         \\
\multirow{-2}{*}{FuzzQE*}   & \small{\checkc{\Checkmark}}   & \textbf{18.4} & \textbf{24.2} & \textbf{17.3} & \cellcolor[HTML]{EFEFEF}43.2 & \cellcolor[HTML]{EFEFEF}13.1 & \cellcolor[HTML]{EFEFEF}10.9 & \cellcolor[HTML]{EFEFEF}33.4 & \cellcolor[HTML]{EFEFEF}47.6 & \cellcolor[HTML]{EFEFEF}25.9 & \cellcolor[HTML]{EFEFEF}18.1 & \cellcolor[HTML]{EFEFEF}14.6 & \cellcolor[HTML]{EFEFEF}10.7 & \cellcolor[HTML]{EFEFEF}8.9 & \cellcolor[HTML]{EFEFEF}11.7 & \cellcolor[HTML]{EFEFEF}7.5 & \cellcolor[HTML]{EFEFEF}5.7 & \cellcolor[HTML]{EFEFEF}6.4 \\
 \midrule
\\ [-2ex]\multicolumn{19}{c}{NELL995} 
\\ [-0.1ex]\midrule \\[-2.4ex]                            & \small{\base{\XSolidBrush}}& -             & 18.6    & 12.5      & 33.0                         & 12.1                         & 9.6                          & 27.5                         & 35.2                         & 18.4                         & 14.4                         & 8.5                          & 8.7                          & -                           & -                           & -                            & -                           & -                           \\
\multirow{-2}{*}{GQE}       & \small{\checkc{\Checkmark}}   & -             & \textbf{21.4} & \textbf{14.6} & \cellcolor[HTML]{EFEFEF}39.4 & \cellcolor[HTML]{EFEFEF}13.2 & \cellcolor[HTML]{EFEFEF}10.8 & \cellcolor[HTML]{EFEFEF}31.7 & \cellcolor[HTML]{EFEFEF}38.9 & \cellcolor[HTML]{EFEFEF}20.8 & \cellcolor[HTML]{EFEFEF}17.5 & \cellcolor[HTML]{EFEFEF}10.3 & \cellcolor[HTML]{EFEFEF}9.6  & \cellcolor[HTML]{EFEFEF}-   & \cellcolor[HTML]{EFEFEF}-   & \cellcolor[HTML]{EFEFEF}-    & \cellcolor[HTML]{EFEFEF}-   & \cellcolor[HTML]{EFEFEF}-   \\
                            & \small{\base{\XSolidBrush}}& -             & 22.9     &15.2     & 42.3                         & 14.0                         & 11.0                         & 33.3                         & 44.6                         & 22.5                         & 16.8                         & 11.3                         & 10.3                         & -                           & -                           & -                            & -                           & -                           \\
\multirow{-2}{*}{Query2Box} & \small{\checkc{\Checkmark}}   & -             & \textbf{25.5} & \textbf{16.5} & \cellcolor[HTML]{EFEFEF}49.7 & \cellcolor[HTML]{EFEFEF}15.3 & \cellcolor[HTML]{EFEFEF}12.2 & \cellcolor[HTML]{EFEFEF}37.7 & \cellcolor[HTML]{EFEFEF}48.7 & \cellcolor[HTML]{EFEFEF}24.2 & \cellcolor[HTML]{EFEFEF}18.5 & \cellcolor[HTML]{EFEFEF}11.9 & \cellcolor[HTML]{EFEFEF}11.4 & \cellcolor[HTML]{EFEFEF}-   & \cellcolor[HTML]{EFEFEF}-   & \cellcolor[HTML]{EFEFEF}-    & \cellcolor[HTML]{EFEFEF}-   & \cellcolor[HTML]{EFEFEF}-   \\
                            & \small{\base{\XSolidBrush}}& -             & 32.6     & 22.5     & 62.7                         & 18.2                         & 14.9                         & 48.5                         & 58.6                         & 33.4                         & 23.1                         & 18.7                         & 14.9                         & -                           & -                           & -                            & -                           & -                           \\
\multirow{-2}{*}{ABIN}      & \small{\checkc{\Checkmark}}   & -             & \textbf{34.0} &\textbf{23.8} & \cellcolor[HTML]{EFEFEF}66.1 & \cellcolor[HTML]{EFEFEF}19.0 & \cellcolor[HTML]{EFEFEF}15.2 & \cellcolor[HTML]{EFEFEF}50.2 & \cellcolor[HTML]{EFEFEF}60.5 & \cellcolor[HTML]{EFEFEF}34.7 & \cellcolor[HTML]{EFEFEF}24.4 & \cellcolor[HTML]{EFEFEF}19.8 & \cellcolor[HTML]{EFEFEF}16.2 & \cellcolor[HTML]{EFEFEF}-   & \cellcolor[HTML]{EFEFEF}-   & \cellcolor[HTML]{EFEFEF}-    & \cellcolor[HTML]{EFEFEF}-   & \cellcolor[HTML]{EFEFEF}-   \\
                            & \small{\base{\XSolidBrush}}& 18.3          & 24.6      & 14.8    & 52.8                         & 13.1                         & 11.4                         & 37.5                         & 47.6                         & 24.2                         & 14.3                         & 12.2                         & 8.6                          & 5.1                         & 7.8                         & 11.6                         & 4.6                         & 5.4                         \\
\multirow{-2}{*}{BetaE}     & \small{\checkc{\Checkmark}}   & \textbf{20.2} & \textbf{27.4} &\textbf{17.3} & \cellcolor[HTML]{EFEFEF}58.9 & \cellcolor[HTML]{EFEFEF}16.0 & \cellcolor[HTML]{EFEFEF}13.5 & \cellcolor[HTML]{EFEFEF}40.2 & \cellcolor[HTML]{EFEFEF}49.6 & \cellcolor[HTML]{EFEFEF}27.9 & \cellcolor[HTML]{EFEFEF}17.0 & \cellcolor[HTML]{EFEFEF}14.8 & \cellcolor[HTML]{EFEFEF}9.3  & \cellcolor[HTML]{EFEFEF}5.7 & \cellcolor[HTML]{EFEFEF}8.1 & \cellcolor[HTML]{EFEFEF}11.9 & \cellcolor[HTML]{EFEFEF}5.0 & \cellcolor[HTML]{EFEFEF}5.5 \\
                            & \small{\base{\XSolidBrush}}& 20.0          & 27.3     &18.3     & 55.9                         & 16.8                         & 14.9                         & 36.4                         & 48.0                         & 26.1                         & 19.7                         & 15.6                         & 11.9                         & 7.1                         & 8.9                         & 10.9                         & 3.5                         & 4.3                         \\
\multirow{-2}{*}{FuzzQE*}   & \small{\checkc{\Checkmark}}   & \textbf{21.4} & \textbf{29.0} & \textbf{19.9} & \cellcolor[HTML]{EFEFEF}59.4 & \cellcolor[HTML]{EFEFEF}17.2 & \cellcolor[HTML]{EFEFEF}15.4 & \cellcolor[HTML]{EFEFEF}39.2 & \cellcolor[HTML]{EFEFEF}50.2 & \cellcolor[HTML]{EFEFEF}27.7 & \cellcolor[HTML]{EFEFEF}21.4 & \cellcolor[HTML]{EFEFEF}16.8 & \cellcolor[HTML]{EFEFEF}13.6 & \cellcolor[HTML]{EFEFEF}8.7 & \cellcolor[HTML]{EFEFEF}9.5 & \cellcolor[HTML]{EFEFEF}11.2 & \cellcolor[HTML]{EFEFEF}4.0 & \cellcolor[HTML]{EFEFEF}5.1 \\ \midrule
\\ [-2ex]\multicolumn{19}{c}{WN18} 
\\ [-0.1ex]\midrule \\[-2.4ex]                            & \small{\base{\XSolidBrush}}& -             & 28.4    &19.6      & 56.5                         & 15.8                         & 11.3                         & 39.0                         & 54.7                         & 26.9                         & 19.4                         & 21.3                         & 10.6                         & -                            & -                            & -                            & -                            & -                            \\
\multirow{-2}{*}{GQE}       & \small{\checkc{\Checkmark}}   & -             & \textbf{30.9} & \textbf{21.1} & \cellcolor[HTML]{EFEFEF}59.4 & \cellcolor[HTML]{EFEFEF}18.2 & \cellcolor[HTML]{EFEFEF}13.0 & \cellcolor[HTML]{EFEFEF}44.2 & \cellcolor[HTML]{EFEFEF}59.0 & \cellcolor[HTML]{EFEFEF}28.4 & \cellcolor[HTML]{EFEFEF}21.2 & \cellcolor[HTML]{EFEFEF}23.0 & \cellcolor[HTML]{EFEFEF}11.9 & \cellcolor[HTML]{EFEFEF}-    & \cellcolor[HTML]{EFEFEF}-    & \cellcolor[HTML]{EFEFEF}-    & \cellcolor[HTML]{EFEFEF}-    & \cellcolor[HTML]{EFEFEF}-    \\
                            & \small{\base{\XSolidBrush}}& -             & 40.1     & 30.8     & 70.2                         & 26.2                         & 15.3                         & 56.6                         & 68.9                         & 42.4                         & 28.9                         & 35.4                         & 16.7                         & -                            & -                            & -                            & -                            & -                            \\
\multirow{-2}{*}{Query2Box} & \small{\checkc{\Checkmark}}   & -             & \textbf{43.2} &\textbf{33.8} & \cellcolor[HTML]{EFEFEF}75.6 & \cellcolor[HTML]{EFEFEF}28.6 & \cellcolor[HTML]{EFEFEF}16.9 & \cellcolor[HTML]{EFEFEF}58.8 & \cellcolor[HTML]{EFEFEF}73.2 & \cellcolor[HTML]{EFEFEF}45.8 & \cellcolor[HTML]{EFEFEF}32.2 & \cellcolor[HTML]{EFEFEF}38.1 & \cellcolor[HTML]{EFEFEF}19.2 & \cellcolor[HTML]{EFEFEF}-    & \cellcolor[HTML]{EFEFEF}-    & \cellcolor[HTML]{EFEFEF}-    & \cellcolor[HTML]{EFEFEF}-    & \cellcolor[HTML]{EFEFEF}-    \\
                            & \small{\base{\XSolidBrush}}& -             & 49.8     &42.6     & 77.1                         & 29.6                         & 26.7                         & 65.2                         & 78.9                         & 52.4                         & 39.7                         & 48.6                         & 29.7                         & -                            & -                            & -                            & -                            & -                            \\
\multirow{-2}{*}{ABIN}      & \small{\checkc{\Checkmark}}   & -             & \textbf{51.1} &\textbf{44.0} & \cellcolor[HTML]{EFEFEF}78.4 & \cellcolor[HTML]{EFEFEF}30.2 & \cellcolor[HTML]{EFEFEF}27.5 & \cellcolor[HTML]{EFEFEF}66.9 & \cellcolor[HTML]{EFEFEF}80.2 & \cellcolor[HTML]{EFEFEF}54.4 & \cellcolor[HTML]{EFEFEF}40.6 & \cellcolor[HTML]{EFEFEF}50.2 & \cellcolor[HTML]{EFEFEF}30.8 & \cellcolor[HTML]{EFEFEF}-    & \cellcolor[HTML]{EFEFEF}-    & \cellcolor[HTML]{EFEFEF}-    & \cellcolor[HTML]{EFEFEF}-    & \cellcolor[HTML]{EFEFEF}-    \\
                            & \small{\base{\XSolidBrush}}& 32.9          & 42.5      &32.6    & 70.7                         & 27.8                         & 25.5                         & 57.5                         & 70.5                         & 43.6                         & 30.4                         & 37.7                         & 18.5                         & 18.8                         & 17.4                         & 15.2                         & 10.2                         & 16.3                         \\
\multirow{-2}{*}{BetaE}     & \small{\checkc{\Checkmark}}   & \textbf{34.4} & \textbf{44.6} & \textbf{35.1} & \cellcolor[HTML]{EFEFEF}74.8 & \cellcolor[HTML]{EFEFEF}28.4 & \cellcolor[HTML]{EFEFEF}26.2 & \cellcolor[HTML]{EFEFEF}59.2 & \cellcolor[HTML]{EFEFEF}72.8 & \cellcolor[HTML]{EFEFEF}46.3 & \cellcolor[HTML]{EFEFEF}32.8 & \cellcolor[HTML]{EFEFEF}40.0 & \cellcolor[HTML]{EFEFEF}21.1 & \cellcolor[HTML]{EFEFEF}19.1 & \cellcolor[HTML]{EFEFEF}17.5 & \cellcolor[HTML]{EFEFEF}15.5 & \cellcolor[HTML]{EFEFEF}10.8 & \cellcolor[HTML]{EFEFEF}16.5 \\
                            & \small{\base{\XSolidBrush}}& 33.1          & 43.9    & 34.6      & 73.9                         & 28.4                         & 26.0                         & 58.2                         & 70.1                         & 45.5                         & 31.8                         & 40.3                         & 20.7                         & 15.9                         & 16.0                         & 14.9                         & 8.7                          & 13.4                         \\
\multirow{-2}{*}{FuzzQE*}   & \small{\checkc{\Checkmark}}   & \textbf{34.6} & \textbf{45.6} &\textbf{36.2} & \cellcolor[HTML]{EFEFEF}76.5 & \cellcolor[HTML]{EFEFEF}29.1 & \cellcolor[HTML]{EFEFEF}26.4 & \cellcolor[HTML]{EFEFEF}60.5 & \cellcolor[HTML]{EFEFEF}73.7 & \cellcolor[HTML]{EFEFEF}47.6 & \cellcolor[HTML]{EFEFEF}33.2 & \cellcolor[HTML]{EFEFEF}41.2 & \cellcolor[HTML]{EFEFEF}22.6 & \cellcolor[HTML]{EFEFEF}16.8 & \cellcolor[HTML]{EFEFEF}16.4 & \cellcolor[HTML]{EFEFEF}15.5 & \cellcolor[HTML]{EFEFEF}9.8  & \cellcolor[HTML]{EFEFEF}15.4 \\ \bottomrule

\end{tabular}}
\label{table:1}
\end{table*} 

\begin{table}[t]
\centering
\setlength\tabcolsep{1pt}
\setlength\abovecaptionskip{2pt}
\caption{Overall MRR results (\%) over four datasets.}
\scalebox{0.95}{
\begin{tabular}{@{\extracolsep{4pt}}lcccccccc@{}}
\toprule[1pt]
\\[-2.9ex]
                    & \multicolumn{2}{c}{FB15k}      & \multicolumn{2}{c}{FB15k-237}  & \multicolumn{2}{c}{NELL995}       & \multicolumn{2}{c}{WN18}       \\[-0.5ex]   \cline{2-3} \cline{4-5} \cline{6-7} \cline{8-9} 
\\[-2.7ex] 
\multirow{-2}{*}{Model} & ${\rm avg}$ &${\rm avg}_p$ & ${\rm avg}$ & ${\rm avg}_p$ & ${\rm avg}$ & ${\rm avg}_p$ & ${\rm avg}$ & ${\rm avg}_p$  \\  [-0.5ex] 
\midrule
GQE       & -             & 28.0   & -             & 16.4   & -             & 18.6   & -             & 28.4   \\
Query2Box & -             & 37.9   & -             & 20.6   & -             & 22.9   & -             & 40.1   \\
ABIN      & -             & 48.1   & -             & 30.2   & -             & 32.6   & -             & 49.8   \\
\textit{LoD-ABIN}  & - &\textbf{49.3} & - & \textbf{32.1} & - & \textbf{34.0} & - & \textbf{51.1} \\\midrule
BetaE     & 31.1   & 41.7   & 15.5   & 21.0   & 18.3   & 24.6   & 32.9   & 42.5   \\
FuzzQE*   & 31.5   & 41.9   & 17.1   & 22.8   & 20.0   & 27.3   & 33.1   & 43.9   \\
\textit{LoD-FuzzQE*}      & \textbf{33.1} & \textbf{43.8} & \textbf{18.3} & \textbf{24.2} & \textbf{21.4} & \textbf{29.0} & \textbf{34.6} & \textbf{45.6}  \\ \bottomrule[1pt] 
\end{tabular}}
\label{table:2}
\end{table}

\section{Experiments}
\label{sec:exp}
\subsection{Evaluation Setup}
\begin{table*}[ht]
\renewcommand{\arraystretch}{0.7}
\centering
\setlength{\abovecaptionskip}{2pt}%
\caption{Ablation MRR results (\%) on FB15k. Model implementation based on Query2box. ${\rm avg}_p$ denotes the average MRR on EPFO. ${\rm avg}_{unseen}$ denotes the the average MRR on $\{{\rm pi}, {\rm ip}, {\rm 2u}, {\rm up}\}$.}
\scalebox{0.95}{
\begin{tabular}{@{}cccccccccccccc@{}}
\toprule[1pt]
\textit{Hier\_conj} & \textit{L\_prompt} & \textit{Grad\_adapt} & ${\rm avg}_{p}$ & ${\rm avg}_{unseen}$ & ${\rm 1p}$ & ${\rm 2p}$ & ${\rm 3p}$ & ${\rm 2i}$ & ${\rm 3i}$ & ${\rm pi}$ & ${\rm ip}$ & ${\rm 2u}$ & ${\rm up}$ \\ \midrule
                            &                           &                           & 37.9 & 29.3 & 67.2 & 21.7 & 14.3 & 54.9 & 66.3 & 39.7 & 25.9 & 34.9 & 16.5 \\
\small{\Checkmark} &                           &                           & 38.6 & 30.9 & 67.1 & 21.5 & 14.2 & 54.8 & 66.5 & 41.5 & 27.7 & 36.7 & 17.8 \\
                            & \small{\Checkmark} &                           & 38.0 & 29.3 & 67.5 & 21.8 & 14.3 & 55.1 & 66.4 & 39.6 & 25.9 & 35.1 & 16.6 \\
                            &                           & \small{\Checkmark} & 38.1 & 29.4 & 67.1 & 21.6 & 14.2 & 55.3 & 66.6 & 39.8 & 26.1 & 35.2 & 16.6 \\
                            & \small{\Checkmark} & \small{\Checkmark} & 38.2 & 29.5 & 67.4 & 21.8 & 14.3 & 55.4 & 66.7 & 39.8 & 26.1 & 35.3 & 16.7 \\
\small{\Checkmark} &                           & \small{\Checkmark} & 38.7 & 31.0 & 67.0 & 21.5 & 14.2 & 55.0 & 66.5 & 41.4 & \textbf{28.1} & 36.6 & 17.8 \\
\small{\Checkmark} & \small{\Checkmark} &                           & 38.7 & 30.8 & 67.3 & 21.8 & 14.3 & 55.2 & 66.6 & 41.3 & 27.7 & 36.4 & 17.6 \\
\small{\Checkmark} & \small{\Checkmark} & \small{\Checkmark} & \textbf{39.0} & \textbf{31.1} & \textbf{67.7} & \textbf{22.0} & \textbf{14.4} & \textbf{55.5} & \textbf{66.9} & \textbf{41.7} & 27.8 & \textbf{36.8} & \textbf{17.9}\\ \bottomrule[1pt]
\end{tabular}
}
\label{table:4}
\end{table*}
%

%

\noindent\textbf{Datasets.}
We evaluate our approach over 4 standard KG datasets including \textbf{FB15k} \cite{TransE&FB15k&WN18}, \textbf{FB15k-237} \cite{FB15k-237}, \textbf{NELL995} \cite{NELL995} and \textbf{WN18} \cite{TransE&FB15k&WN18} with their with official training / validation / test edge splits shown in Table~\ref{table:a} containing 14 types of queries, which are created by the query construction method in \cite{BetaE&KGreasoning}. Following \cite{BetaE&KGreasoning}, we train models which can not handle negation with queries of ${\rm 1p, 2p, 3p, 2i, 3i}$ patterns and evaluate those with all EPFO. While FOL models over atomic queries plus ${\rm 2ni, 3ni, inp, pni, pin}$, and evaluate those over all FOL patterns. Note that BetaE \cite{BetaE&KGreasoning} generates queries with answers less than a threshold, while GQE \cite{GQE} and Query2Box \cite{Q2B} do not limit answers. We evaluate these methods following BetaE for fair comparison. Besides, as for constructing queries of noisy knowledge graphs, we randomly mix training / validation / test of FB15k and NELL995 up with different proportions respectively.
%

\noindent\textbf{Evaluation Metrics.}
Following \cite{BetaE&KGreasoning}, for each non-trivial answer $t$ of test query $q$, we rank it against non-answer entities $\mathcal{E} \backslash [\![q]\!]_{\rm test}$ \cite{TransE&FB15k&WN18}. Then the rank of each answer is labeled as $r$. We use \textbf{Mean Reciprocal Rank(MRR)}: $\frac{1}{r}$ and \textbf{Hits at} $\mathbf{N \ ({\rm \textbf{H}}@N)}: 1[r \leq N]$ as quantitative metrics.
%

\noindent\textbf{Baselines.}
We consider five metric learning based KG reasoning models over incomplete KG: \textbf{GQE} \cite{GQE}, \textbf{Query2Box} \cite{Q2B}, \textbf{ABIN} \cite{ABIN}, \textbf{BetaE} \cite{BetaE&KGreasoning} and \textbf{FuzzQE} \cite{FuzzQE}. Note that GQE, Query2Box and ABIN cannot handle the negation operation, so we only evaluate these three methods by EPFO which does not contain the negation operation compared to FOL.
All comparison methods are reproduced by the released source code except FuzzQE. 
As reported in the GitHub project, it usually takes four days to a week to finish a run, which is much more time consuming than other comparison methods. Limited by our experimental conditions and in order to be comparable, we manually reproduce the Logical Mapping Function of FuzzQE as it was described in the paper and marked as FuzzQE* in Table~\ref{table:1} and Table~\ref{table:2}.
%

\noindent\textbf{Implementation Details.}
At the very beginning, all nodes and relationships in the knowledge graph need to be embedded onto feature space in order to participate in model training. So in order to obtain better representations of nodes and relations, we pre-train the embedding model (i.e., AcrE) and then reload parameters of embedding layer into our end-to-end process as initialization parameters. 
Specifically, AcrE uses $\tau$ to denote a concatenating operation and 2-dimensional (2D) reshaping function. 
\begin{equation}
    {\rm C}^0 = \omega^0_c \star \tau({\rm [e;r]}) + {\rm b}^0_c
\end{equation} 
\begin{equation}
    {\rm C}^l = \omega^l_c\star {\rm C}^{l-1} + {\rm b}^l_c
\end{equation}
\begin{equation}
    o  = Flatten({\rm ReLU}({\rm C}^L+\tau({\rm [e;r]})))
\end{equation}
where $\star$ denotes a 2D convolution operation, $\omega^0_c$ is a standard filter while $\omega^l_c$ is a dilated filter, ${\rm b}^0_c$ and ${\rm b}^l_c$ are bias vectors. ${\rm C}^l$ denotes the output after $l$ convolutions while ${\rm C}^L$ is the output of the last dilated convolution. We train the model by optimizing a listwise loss function for 
 20 epochs and take 512 as the embedding dimension $d$ of nodes.

Besides, inspired by the theory of Distributional Hypothesis \cite{distributional},
\cite{neighborhood,redGCNneib2,neib1,neib2} proposes that aggregating local information to augment entity representation is helpful for KGR tasks. Therefore we search the neighbor entity set of each entity contained in reasoning sub-graphs for once, and align the maximum of neighbors as 64. Following~\cite{ConvE}, we concatenate the embedding of the center entity and neighbors and then do feature fusion by a 2D standard convolution. Suppose an entity ${\rm e} \in \mathbbm{R}^d $ and its aligned neighbor set ${N_{\rm e}^\prime}$, the new representation ${\rm e}^\prime = {\rm MLP}(Flatten({\rm ReLU}(\omega_n \star ({\tau}^\prime ({\rm e}, {N_{\rm e}}^\prime)) + {\rm b}_n))$, where $\star$ denotes a 2D convolution operation, $\omega_n$ is the filter, ${\rm b}_n $ is the bias and the specification of ${\rm MLP}$ is $\mathbbm{R}^{m_1 \times m_2} \times \mathbbm{R}^d$. We define the concatenate function ${\tau}^\prime ({\rm e}, N_{\rm e}) \in \mathbbm{R}^{m_1 \times m_2}$ as $[{\rm e}; {{\rm e}_{neib}}^{1^\prime} ; {{\rm e}_{neib}}^{2^\prime}; \ldots ;{{\rm e}_{neib}}^m]$ where ${{\rm e}_{neib}}^i \in {N_{\rm e}^\prime}$.

\noindent\textbf{Hyparameters.}
We adjust the following hyperparameters to obtain the best model performance. Noted that we use the same hyperparameters before and after applying \textit{LoD} for each base model.
\begin{itemize}
    \item Dimensions to the embeddings $d$ from \{256, 378, 512, 768, 1024\} 
    \item Learning rate ${\rm lr}$ from \{1e$-$4, 5e$-$3, 1e$-$3\}
    \item Batch size $b_z$ from \{128, 256, 512\}
    \item Negative sample size $n\_s$ from \{32, 64, 128\}
    \item Maximum of neighbors $M$ from \{0, 16, 32, 64, 128\}
    \item Length of Logic-specific prompt $d_T$ from \{8, 16, 32, 64\}
    \item Attenuation Factor $z$ is from \{0.10, 0.05, 0.01\}
\end{itemize}
The best set of \textit{LoD} hyperparameters is shown in Table~\ref{table:ap2}.

\subsection{Performance}
\noindent\textbf{Overall.}
As shown in Table~\ref{table:1}, we apply \textit{LoD} on several mainstream models on FB15k. Table~\ref{table:2} shows the comparative experiments of best implementation of \textit{LoD} on four datasets. Note that to implement \textit{LoD} using the same backbone, we evaluate EPFO (i.e., only ${\rm avg}_p$) and FOL (i.e., ${\rm avg}$ and ${\rm avg}_p$) in the top and bottom halves of Table~\ref{table:2} respectively.
For EPFO, we implement \textit{LoD} based on ABIN (noted as \textit{LoD-ABIN}), while for FOL, by FuzzQE* (noted as \textit{LoD-FuzzQE*}). It shows that \textit{LoD-ABIN} achieves gains of \textbf{1.2}\%, \textbf{1.9}\%, \textbf{1.4}\% and \textbf{1.3}\% in ${\rm avg}_p$ MRR, \textit{LoD-FuzzQE*} achieves \textbf{1.6}\%, \textbf{1.2}\%, \textbf{1.4}\% and \textbf{1.5}\% in ${\rm avg}$ MRR on FB15k, FB15k-237, NELL995 and WN18 respectively. Apparently, \textit{LoD} benefits from both representation enhancement and logic augmentation by aggregating neighboring information.
\begin{table*}[ht]
\centering
\setlength{\abovecaptionskip}{2pt}%
\caption{MRR results (\%) on noisy knowledge graph with 80\% FB15k and 20\% NELL995. Methods with \textbf{\textit{\emphasize{+dl}}} indicates using \textit{LoD Loss}. ${\rm avg}_p$ indicates the the average MRR on EPFO. ${\rm \textbf{Decline}}$ indicates the MRR drop compared to without noise. ${\rm \textbf{Recovery}}$ indicates the MRR recovery by \textit{LoD Loss}. (i.e., 7.7-2.0 = 5.7)}
\scalebox{0.95}{
\begin{tabular}{@{}clcccccccccccc@{}}
\toprule[1pt]
\\[-3.0ex]
\begin{tabular}[c]{@{}c@{}}Setting\\ Up\end{tabular} & Model                             &   \begin{tabular}[c]{@{}c@{}}${\rm avg}_p$ \\ ${\rm \textbf{Recovery}}$ \end{tabular} $\uparrow$  & \begin{tabular}[c]{@{}c@{}}${\rm avg}_p$ \\ ${\rm \textbf{Decline}}$ \end{tabular} $\downarrow$   &${\rm avg}_p$ $\uparrow$              & ${\rm 1p}$                   & ${\rm 2p}$                   & ${\rm 3p}$                   & ${\rm 2i}$                   & ${\rm 3i}$                   & ${\rm pi}$                   & ${\rm ip}$                   & ${\rm 2u}$                   & ${\rm up}$                   \\[-0.3ex] \midrule
                                                     & GQE                               & -                            & 7.7                         & 20.3                         & 33.8                         & 13.0                         & 10.1                         & 29.7                         & 39.9                         & 19.8                         & 16.3                         & 11.0                         & 9.3                          \\
                                                     & \cellcolor[HTML]{EFEFEF}GQE \textbf{\textit{\emphasize{+dl}}}       & \cellcolor[HTML]{EFEFEF}5.7  & \cellcolor[HTML]{EFEFEF}2.0 & \cellcolor[HTML]{EFEFEF}26.0 & \cellcolor[HTML]{EFEFEF}49.4 & \cellcolor[HTML]{EFEFEF}13.9 & \cellcolor[HTML]{EFEFEF}10.7 & \cellcolor[HTML]{EFEFEF}36.7 & \cellcolor[HTML]{EFEFEF}48.6 & \cellcolor[HTML]{EFEFEF}25.4 & \cellcolor[HTML]{EFEFEF}17.9 & \cellcolor[HTML]{EFEFEF}20.3 & \cellcolor[HTML]{EFEFEF}10.7 \\
                                                     & Query2Box                         & -                            & 12.1                        & 25.9                         & 45.7                         & 15.8                         & 12.2                         & 37.5                         & 49.8                         & 23.4                         & 18.2                         & 18.7                         & 11.6                         \\
\multirow{-4}{*}{Setting 1}                          & \cellcolor[HTML]{EFEFEF}Query2Box \textbf{\textit{\emphasize{+dl}}} & \cellcolor[HTML]{EFEFEF}9.8  & \cellcolor[HTML]{EFEFEF}2.3 & \cellcolor[HTML]{EFEFEF}35.7 & \cellcolor[HTML]{EFEFEF}64.6 & \cellcolor[HTML]{EFEFEF}20.4 & \cellcolor[HTML]{EFEFEF}13.6 & \cellcolor[HTML]{EFEFEF}51.4 & \cellcolor[HTML]{EFEFEF}64.5 & \cellcolor[HTML]{EFEFEF}37.2 & \cellcolor[HTML]{EFEFEF}23.6 & \cellcolor[HTML]{EFEFEF}30.7 & \cellcolor[HTML]{EFEFEF}14.9 \\ \midrule
                                                     & GQE                              & -                            & 11.8                        & 16.2                         & 30.4                         & 10.7                         & 8.4                          & 21.3                         & 30.5                         & 16.1                         & 12.5                         & 7.9                          & 8.0                          \\
                                                     & \cellcolor[HTML]{EFEFEF}GQE \textbf{\textit{\emphasize{+dl}}}     & \cellcolor[HTML]{EFEFEF}6.7  & \cellcolor[HTML]{EFEFEF}5.1 & \cellcolor[HTML]{EFEFEF}22.9 & \cellcolor[HTML]{EFEFEF}51.2 & \cellcolor[HTML]{EFEFEF}14.4 & \cellcolor[HTML]{EFEFEF}10.5 & \cellcolor[HTML]{EFEFEF}36.4 & \cellcolor[HTML]{EFEFEF}18.3 & \cellcolor[HTML]{EFEFEF}25.8 & \cellcolor[HTML]{EFEFEF}18.1 & \cellcolor[HTML]{EFEFEF}20.5 & \cellcolor[HTML]{EFEFEF}10.8 \\
                                                     & Query2Box                          & -                            & 14.4                        & 23.5                         & 40.6                         & 13.4                         & 11.6                         & 34.8                         & 46.2                         & 22.1                         & 16.4                         & 16.5                         & 10.2                         \\
\multirow{-4}{*}{Setting 2}                          & \cellcolor[HTML]{EFEFEF}Query2Box \textbf{\textit{\emphasize{+dl}}}                        & \cellcolor[HTML]{EFEFEF}11.4 & \cellcolor[HTML]{EFEFEF}3.0 & \cellcolor[HTML]{EFEFEF}34.9 & \cellcolor[HTML]{EFEFEF}64.4 & \cellcolor[HTML]{EFEFEF}19.9 & \cellcolor[HTML]{EFEFEF}13.2 & \cellcolor[HTML]{EFEFEF}51.3 & \cellcolor[HTML]{EFEFEF}62.4 & \cellcolor[HTML]{EFEFEF}36.4 & \cellcolor[HTML]{EFEFEF}22.3 & \cellcolor[HTML]{EFEFEF}29.5 & \cellcolor[HTML]{EFEFEF}14.7 \\ \bottomrule[1pt]
\end{tabular}}
\label{table:5}
\end{table*} 

\begin{table}[ht]
\centering
\setlength\tabcolsep{1pt}
\setlength{\abovecaptionskip}{2pt}
\caption{MRR results (\%) on noisy knowledge graph. Methods with \textbf{\textit{\emphasize{+dl}}} denotes 
those trained with \textit{LoD Loss}. Here ${\rm avg}_p$ denotes the the average MRR on EPFO.}
\scalebox{0.95}{
\begin{tabular}{@{}clcccccc@{}}
\toprule[1pt]
\\[-2.8ex]
    &          & \multicolumn{3}{c}{Setting 1}                                                                 & \multicolumn{3}{c}{Setting 2}                                                                 \\[-0.2ex] \cmidrule(r){3-5} \cmidrule(r){6-8} \specialrule{0em}{1pt}{1pt}
                                \\[-3.3ex] 
\multirow{-2}{*}{$\rho$} & \multirow{-2}{*}{Model}                                        &  $\textbf{Rec.}\uparrow$ &  $\textbf{Dec.}\downarrow$ & ${\rm avg}_p$ $\uparrow$               &  $\textbf{Rec.}\uparrow$ &  $\textbf{Dec.}\downarrow$ & ${\rm avg}_p$ $\uparrow$                \\ [-0.8pt]
\midrule
                                & GQE                                                            & -    & 4.5  & 23.5 & -    & 8.9  & 19.1 \\
                                & \cellcolor[HTML]{EFEFEF}GQE \textbf{\textit{\emphasize{+dl}}}      &  
\cellcolor[HTML]{EFEFEF}2.9  & \cellcolor[HTML]{EFEFEF}1.6  & \cellcolor[HTML]{EFEFEF}26.4 & \cellcolor[HTML]{EFEFEF}5.0    & \cellcolor[HTML]{EFEFEF}3.9  & \cellcolor[HTML]{EFEFEF}24.1 \\& Query2Box                                                      & -    & 5.3  & 32.6 & -    & 12.5 & 25.4 \\
\multirow{-4}{*}{10 }            & \cellcolor[HTML]{EFEFEF}Query2Box \textbf{\textit{\emphasize{+dl}}} &  
\cellcolor[HTML]{EFEFEF}4.1  & \cellcolor[HTML]{EFEFEF}1.2  & \cellcolor[HTML]{EFEFEF}36.7 & \cellcolor[HTML]{EFEFEF}9.9  & \cellcolor[HTML]{EFEFEF}2.6  & \cellcolor[HTML]{EFEFEF}35.3 \\
\midrule
                                & GQE                                                            & -    & 7.7  & 20.3 & -    & 11.8 & 16.2 \\
                                & \cellcolor[HTML]{EFEFEF}GQE \textbf{\textit{\emphasize{+dl}}}      &  
\cellcolor[HTML]{EFEFEF}5.6  & \cellcolor[HTML]{EFEFEF}2.0  & \cellcolor[HTML]{EFEFEF}26.0 & \cellcolor[HTML]{EFEFEF}6.7  & \cellcolor[HTML]{EFEFEF}5.1  & \cellcolor[HTML]{EFEFEF}22.9 \\
                                & Query2Box                                                      & -    & 12.1 & 25.9 & -    & 14.4 & 23.5 \\
\multirow{-4}{*}{20 }            & \cellcolor[HTML]{EFEFEF}Query2Box \textbf{\textit{\emphasize{+dl}}} &  
\cellcolor[HTML]{EFEFEF}9.8  & \cellcolor[HTML]{EFEFEF}2.3  & \cellcolor[HTML]{EFEFEF}35.7 & \cellcolor[HTML]{EFEFEF}11.4 & \cellcolor[HTML]{EFEFEF}3.0  & \cellcolor[HTML]{EFEFEF}34.9 \\
\midrule
                                & GQE                                                            & -    & 10.7 & 17.3 & -    & 13.8 & 14.2 \\
                                & \cellcolor[HTML]{EFEFEF}GQE \textbf{\textit{\emphasize{+dl}}}     &  
\cellcolor[HTML]{EFEFEF}6.9  & \cellcolor[HTML]{EFEFEF}3.8  & \cellcolor[HTML]{EFEFEF}24.2 & \cellcolor[HTML]{EFEFEF}7.1  & \cellcolor[HTML]{EFEFEF}6.7  & \cellcolor[HTML]{EFEFEF}21.3 \\
                                & Query2Box                                                      & -    & 16.3 & 21.6 & -    & 17.8 & 20.1 \\
\multirow{-4}{*}{30 }            & \cellcolor[HTML]{EFEFEF}Query2Box \textbf{\textit{\emphasize{+dl}}} &  
\cellcolor[HTML]{EFEFEF}12.9 & \cellcolor[HTML]{EFEFEF}3.4  & \cellcolor[HTML]{EFEFEF}34.5 & \cellcolor[HTML]{EFEFEF}13.2 & \cellcolor[HTML]{EFEFEF}4.6  & \cellcolor[HTML]{EFEFEF}33.3 \\
\midrule
                                & GQE                                                            & -    & 12.5 & 15.5 & -    & 15.7 & 12.3 \\
                                & \cellcolor[HTML]{EFEFEF}GQE \textbf{\textit{\emphasize{+dl}}}      &  
\cellcolor[HTML]{EFEFEF}4.9  & \cellcolor[HTML]{EFEFEF}7.6  & \cellcolor[HTML]{EFEFEF}20.4 & \cellcolor[HTML]{EFEFEF}8.3  & \cellcolor[HTML]{EFEFEF}7.4  & \cellcolor[HTML]{EFEFEF}20.6 \\
                                & Query2Box                                                      & -    & 18.3 & 19.6 & -    & 19.5 & 18.4 \\
\multirow{-4}{*}{40 }            & \cellcolor[HTML]{EFEFEF}Query2Box \textbf{\textit{\emphasize{+dl}}} &  
\cellcolor[HTML]{EFEFEF}10.3 & \cellcolor[HTML]{EFEFEF}8.0  & \cellcolor[HTML]{EFEFEF}29.9 & \cellcolor[HTML]{EFEFEF}13.4 & \cellcolor[HTML]{EFEFEF}6.1  & \cellcolor[HTML]{EFEFEF}31.8 \\
\midrule
                                & GQE                                                            & -    & 14.6 & 13.4 & -    & 17.1 & 10.9 \\
                                & \cellcolor[HTML]{EFEFEF}GQE \textbf{\textit{\emphasize{+dl}}}     &  
\cellcolor[HTML]{EFEFEF}2.8  & \cellcolor[HTML]{EFEFEF}11.8 & \cellcolor[HTML]{EFEFEF}16.2 & \cellcolor[HTML]{EFEFEF}9.3  & \cellcolor[HTML]{EFEFEF}7.8  & \cellcolor[HTML]{EFEFEF}20.2 \\
                                & Query2Box                                                      & -    & 20.6 & 17.3 & -    & 22.3 & 15.6 \\
\multirow{-4}{*}{50 }            & \cellcolor[HTML]{EFEFEF}Query2Box \textbf{\textit{\emphasize{+dl}}} &  
\cellcolor[HTML]{EFEFEF}7.3  & \cellcolor[HTML]{EFEFEF}13.3 & \cellcolor[HTML]{EFEFEF}24.6 & \cellcolor[HTML]{EFEFEF}13.8 & \cellcolor[HTML]{EFEFEF}8.5  & \cellcolor[HTML]{EFEFEF}29.4 \\ \bottomrule[1pt]

\end{tabular}}
\label{table:6}
\end{table}

\noindent\textbf{Ablation on LoD Architecture.}
In this section, we perform ablation experiments of each part in LoD Architecture. To highlight the impact and to be fair, we remove the KGE pre-training and neighboring feature fusion in these experiments. As shown in Tabel~\ref{table:4}, \textit{Hier\_conj} means Hierarchical Conjunctive Query, \textit{L\_prompt} means Logic-specific Prompt with $d_T = 32$ and \textit{Grad\_adapt} means Gradient Adaption with $z=0.05$. The results show that \textit{Hier\_conj}, \textit{L\_prompt} and \textit{Grad\_adapt} achieves gains of \textbf{0.7}\%, \textbf{0.1}\% and \textbf{0.2}\% in ${\rm avg}_p$ MRR respectively. \textit{Hier\_conj} achieves gains of \textbf{1.6}\% in ${\rm avg}_{unseen}$ MRR. Activating all of sub-modules can achieve the final gain of \textbf{1.1}\% in ${\rm avg}_p$ MRR and \textbf{1.8}\% in ${\rm avg}_{unseen}$ MRR.
Such results fully demonstrate the effectiveness of \textit{LoD}. In particular, the performance of \textit{Hier\_conj} in ${\rm avg}_{unseen}$ MRR proves that neighbor diffusion from logic perspective has a great significant improvement on unseen FOL.

\begin{figure}
	\begin{center}
 	\includegraphics[width=1\linewidth]{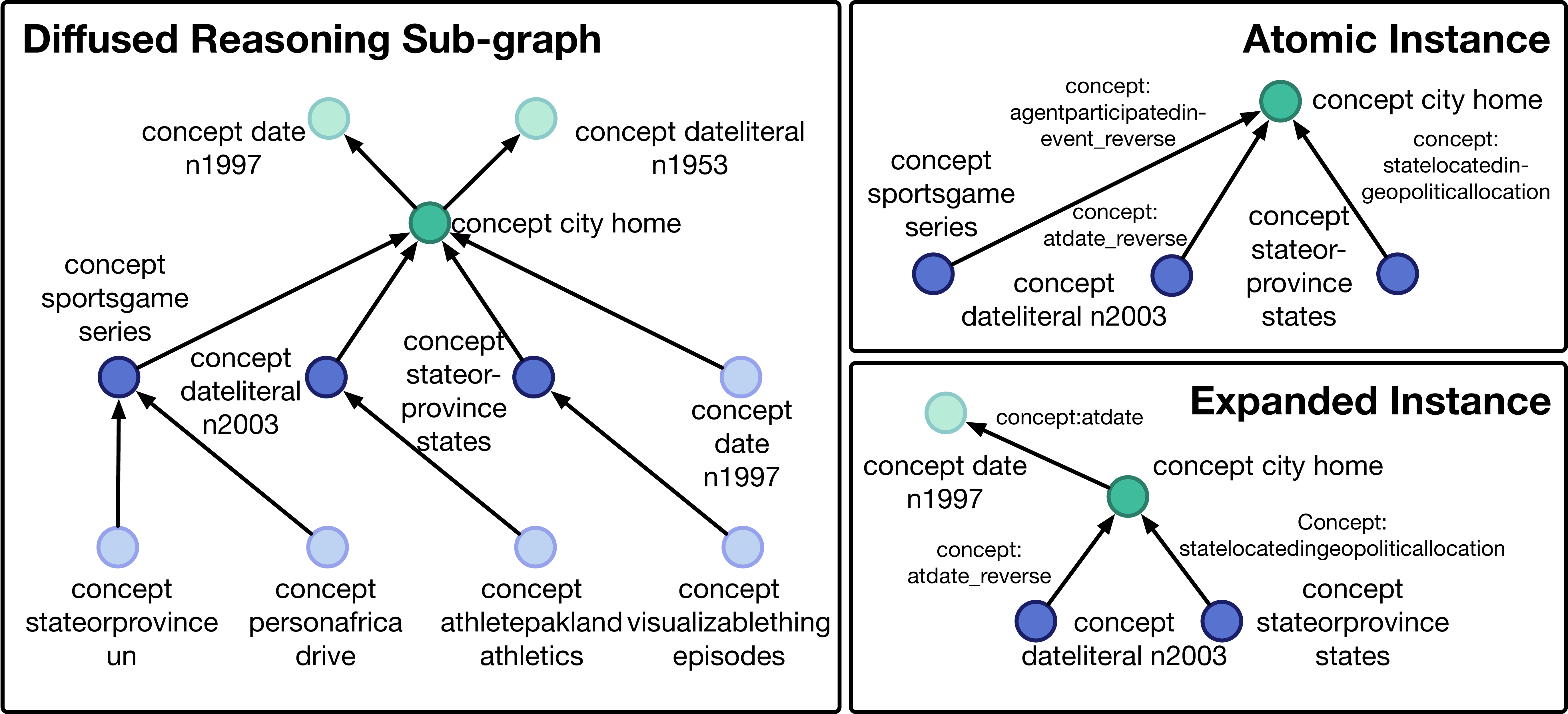}
	\end{center}
	\caption{Illustration of a Logic Diffusion instance of $\rm 3i-ex$ paradigms in NELL995.}
	\label{fig:6}
\end{figure}

\noindent\textbf{Visualization of Logic Diffusion.}
In order to show the results of logic diffusion more intuitively, here we give an example on the NELL995 dataset. As it is illustrated in Figure ~\ref{fig:6}, the atomic training data as input belongs to ${\rm 3i}$ paradigm. By Hierarchical Conjunctive Query, the reasoning sub-graph is diffused as the left. Then a new training instance which belongs to unseen ${\rm ip}$ paradigm is sampled and participate in following training process by random walking. Combining with results in Table~\ref{table:2}, additional training data pre-built by the same way contributes to generalization to unseen paradigms indeed.
\subsection{Ablation Study with Noisy Data}
\label{sec:exploss}
\noindent\textbf{Overall.}
%
As shown in Tabel~\ref{table:5}, we evaluate our \textit{LoD Loss} on noisy knowledge graph which consists of 80\% FB15k as data of rigorous logic and 20\% NELL995 as that of non-rigorous logic. Note that GQE uses a max-margin loss rather than a contrast loss where a cosine similarity is used as the similarity metric. In retrospect, we divide the noisy knowledge graph into (1) training with noise, and (2) testing with noise. Therefore, we 
design two settings:
\begin{itemize}
	\item{Setting 1}: Distinguishing and blocking noisy data during testing, blocking delayed 50,000 steps during training. In this setting, noisy data both in train and test set.
	\item{Setting 2}: Distinguishing and blocking noisy data during testing but no sieves are used during training. In this setting, noisy data only in test set.
\end{itemize}
Since GQE and Query2Box can not handle the negation operation, we only calculate results over EPFO. As shown in Tabel~\ref{table:5}, \textbf{\textit{\emphasize{+dl}}} achieves a gain of \textbf{5.7}\% ${\rm avg}_p$ MRR in setting 1. \textbf{\textit{\emphasize{+dl}}} achieves a gain of \textbf{6.7}\%  ${\rm avg}_p$ MRR in setting 2. These results demonstrate that \textit{LoD Loss} guides models for robust representation and resists the data of non-rigorous logic over noisy knowledgraph during both training and testing.

\noindent\textbf{Noise Ratio.}
Since we take mixed up data of FB15k and NELL995 as noisy knowledge graph, experimentation with different noisy data percentages is also necessary. The results with $(100-\rho$)\% of FB15k and $\rho$\% of NELL995 are shown in Tables \ref{table:6}, where the similarity threshold of GQE is $\varsigma = 0.15$ and Query2Box is $\varsigma = -5.0$.
For setting 1, the proposed \textit{LoD Loss} with Query2Box achieves ${\rm avg}_p$ MRR Recovery of \textbf{4.1}\%, \textbf{9.8}\%, \textbf{12.9}\%, \textbf{10.3}\% and \textbf{7.3}\% when 10\%, 20\%, 30\%, 40\% and 50\% of noisy data are used. For settting 2, it gains \textbf{9.9}\%, \textbf{11.4}\%, \textbf{13.2}\%, \textbf{13.4}\% and \textbf{13.8}\%. Similar results can be observed in other base methods. We can draw three conclusions from these results: (1) \textit{LoD Loss} is stable and effective at various non-rigorous logic ratios; (2) From setting 1, with the use of \textit{LoD Loss}, we don't have to clean up the training set to train a usable model; (3) From setting 2, the inference process of our model is better resist data of non-rigorous logic, which is great valuable for practical application.
%

%% file: sections/6_con.tex
\section{Conclusion}
In this paper, we disassemble the open-set knowledge graph into two major problems: unseen FOL and noise-rich data. To address these issues, we propose a universal module called \textit{LoD} which achieves representation enhancement and logic augmentation. \textit{LoD} discovers unseen queries from surroundings and achieves dynamical equilibrium between different FOL patterns. Besides, we propose a loss function named \textit{LoD Loss} to handle noise-rich data. Extensive experiments on four public datasets demonstrate the superiority of mainstream models with the proposed \textit{LoD} module with mainstream knowledge graph reasoning models over state-of-the-art. Experiments on open-set knowledge graph demonstrate the superiority of \textit{LoD Loss}. Overall, we have done a pioneering work and provided a holistic solution that can be specifically used to solve common sense reasoning on open-set knowledge graph. In future work, we will extend our approach to noisy multi-modal knowledge graph which is a more realistic scene.

%% file: sections/7_ref.tex
\bibliography{acmart}
\bibliographystyle{ACM-Reference-Format}

%% file: main.bbl

\begin{thebibliography}{63}


\ifx \showCODEN    \undefined \def \showCODEN     #1{\unskip}     \fi
\ifx \showDOI      \undefined \def \showDOI       #1{#1}\fi
\ifx \showISBNx    \undefined \def \showISBNx     #1{\unskip}     \fi
\ifx \showISBNxiii \undefined \def \showISBNxiii  #1{\unskip}     \fi
\ifx \showISSN     \undefined \def \showISSN      #1{\unskip}     \fi
\ifx \showLCCN     \undefined \def \showLCCN      #1{\unskip}     \fi
\ifx \shownote     \undefined \def \shownote      #1{#1}          \fi
\ifx \showarticletitle \undefined \def \showarticletitle #1{#1}   \fi
\ifx \showURL      \undefined \def \showURL       {\relax}        \fi
\providecommand\bibfield[2]{#2}
\providecommand\bibinfo[2]{#2}
\providecommand\natexlab[1]{#1}
\providecommand\showeprint[2][]{arXiv:#2}

\bibitem[Arakelyan et~al\mbox{.}(2020)]%
        {CQD}
\bibfield{author}{\bibinfo{person}{Erik Arakelyan}, \bibinfo{person}{Daniel
  Daza}, \bibinfo{person}{Pasquale Minervini}, {and} \bibinfo{person}{Michael
  Cochez}.} \bibinfo{year}{2020}\natexlab{}.
\newblock \showarticletitle{Complex query answering with neural link
  predictors}.
\newblock \bibinfo{journal}{\emph{arXiv preprint arXiv:2011.03459}}
  (\bibinfo{year}{2020}).
\newblock


\bibitem[Bai et~al\mbox{.}(2022)]%
        {particles}
\bibfield{author}{\bibinfo{person}{Jiaxin Bai}, \bibinfo{person}{Zihao Wang},
  \bibinfo{person}{Hongming Zhang}, {and} \bibinfo{person}{Yangqiu Song}.}
  \bibinfo{year}{2022}\natexlab{}.
\newblock \showarticletitle{Query2Particles: Knowledge Graph Reasoning with
  Particle Embeddings}.
\newblock \bibinfo{journal}{\emph{arXiv preprint arXiv:2204.12847}}
  (\bibinfo{year}{2022}).
\newblock


\bibitem[Bordes et~al\mbox{.}(2013)]%
        {TransE&FB15k&WN18}
\bibfield{author}{\bibinfo{person}{Antoine Bordes}, \bibinfo{person}{Nicolas
  Usunier}, \bibinfo{person}{Alberto Garcia-Duran}, \bibinfo{person}{Jason
  Weston}, {and} \bibinfo{person}{Oksana Yakhnenko}.}
  \bibinfo{year}{2013}\natexlab{}.
\newblock \showarticletitle{Translating Embeddings for Modeling
  Multi-relational Data}. In \bibinfo{booktitle}{\emph{Advances in Neural
  Information Processing Systems}}, \bibfield{editor}{\bibinfo{person}{C.J.
  Burges}, \bibinfo{person}{L.~Bottou}, \bibinfo{person}{M.~Welling},
  \bibinfo{person}{Z.~Ghahramani}, {and} \bibinfo{person}{K.Q. Weinberger}}
  (Eds.), Vol.~\bibinfo{volume}{26}. \bibinfo{publisher}{Curran Associates,
  Inc.}
\newblock
\urldef\tempurl%
\url{https://proceedings.neurips.cc/paper/2013/file/1cecc7a77928ca8133fa24680a88d2f9-Paper.pdf}
\showURL{%
\tempurl}


\bibitem[Brown et~al\mbox{.}(2020)]%
        {Brown:LM-fS-learners}
\bibfield{author}{\bibinfo{person}{Tom~B. Brown}, \bibinfo{person}{Benjamin
  Mann}, \bibinfo{person}{Nick Ryder}, \bibinfo{person}{Melanie Subbiah},
  \bibinfo{person}{Jared Kaplan}, \bibinfo{person}{Prafulla Dhariwal},
  \bibinfo{person}{Arvind Neelakantan}, \bibinfo{person}{Pranav Shyam},
  \bibinfo{person}{Girish Sastry}, \bibinfo{person}{Amanda Askell},
  \bibinfo{person}{Sandhini Agarwal}, \bibinfo{person}{Ariel Herbert{-}Voss},
  \bibinfo{person}{Gretchen Krueger}, \bibinfo{person}{Tom Henighan},
  \bibinfo{person}{Rewon Child}, \bibinfo{person}{Aditya Ramesh},
  \bibinfo{person}{Daniel~M. Ziegler}, \bibinfo{person}{Jeffrey Wu},
  \bibinfo{person}{Clemens Winter}, \bibinfo{person}{Christopher Hesse},
  \bibinfo{person}{Mark Chen}, \bibinfo{person}{Eric Sigler},
  \bibinfo{person}{Mateusz Litwin}, \bibinfo{person}{Scott Gray},
  \bibinfo{person}{Benjamin Chess}, \bibinfo{person}{Jack Clark},
  \bibinfo{person}{Christopher Berner}, \bibinfo{person}{Sam McCandlish},
  \bibinfo{person}{Alec Radford}, \bibinfo{person}{Ilya Sutskever}, {and}
  \bibinfo{person}{Dario Amodei}.} \bibinfo{year}{2020}\natexlab{}.
\newblock \showarticletitle{Language Models are Few-Shot Learners}. In
  \bibinfo{booktitle}{\emph{Adv. Neural Inform. Process. Syst.}}
\newblock


\bibitem[Cao et~al\mbox{.}(2021)]%
        {emb4}
\bibfield{author}{\bibinfo{person}{Zongsheng Cao}, \bibinfo{person}{Qianqian
  Xu}, \bibinfo{person}{Zhiyong Yang}, \bibinfo{person}{Xiaochun Cao}, {and}
  \bibinfo{person}{Qingming Huang}.} \bibinfo{year}{2021}\natexlab{}.
\newblock \showarticletitle{Dual quaternion knowledge graph embeddings}. In
  \bibinfo{booktitle}{\emph{Proceedings of the AAAI Conference on Artificial
  Intelligence}}, Vol.~\bibinfo{volume}{35}. \bibinfo{pages}{6894--6902}.
\newblock


\bibitem[Chami et~al\mbox{.}(2020)]%
        {emb2}
\bibfield{author}{\bibinfo{person}{Ines Chami}, \bibinfo{person}{Adva Wolf},
  \bibinfo{person}{Da-Cheng Juan}, \bibinfo{person}{Frederic Sala},
  \bibinfo{person}{Sujith Ravi}, {and} \bibinfo{person}{Christopher R{\'e}}.}
  \bibinfo{year}{2020}\natexlab{}.
\newblock \showarticletitle{Low-dimensional hyperbolic knowledge graph
  embeddings}.
\newblock \bibinfo{journal}{\emph{arXiv preprint arXiv:2005.00545}}
  (\bibinfo{year}{2020}).
\newblock


\bibitem[Chen et~al\mbox{.}(2022a)]%
        {chen2022rlpath}
\bibfield{author}{\bibinfo{person}{Ling Chen}, \bibinfo{person}{Jun Cui},
  \bibinfo{person}{Xing Tang}, \bibinfo{person}{Yuntao Qian},
  \bibinfo{person}{Yansheng Li}, {and} \bibinfo{person}{Yongjun Zhang}.}
  \bibinfo{year}{2022}\natexlab{a}.
\newblock \showarticletitle{RLPath: a knowledge graph link prediction method
  using reinforcement learning based attentive relation path searching and
  representation learning}.
\newblock \bibinfo{journal}{\emph{Applied Intelligence}} \bibinfo{volume}{52},
  \bibinfo{number}{4} (\bibinfo{year}{2022}), \bibinfo{pages}{4715--4726}.
\newblock


\bibitem[Chen et~al\mbox{.}(2018)]%
        {noise}
\bibfield{author}{\bibinfo{person}{Wenhu Chen}, \bibinfo{person}{Wenhan Xiong},
  \bibinfo{person}{Xifeng Yan}, {and} \bibinfo{person}{William~Yang Wang}.}
  \bibinfo{year}{2018}\natexlab{}.
\newblock \showarticletitle{Variational Knowledge Graph Reasoning}. In
  \bibinfo{booktitle}{\emph{NAACL}}.
\newblock


\bibitem[Chen et~al\mbox{.}(2021)]%
        {neighborhood}
\bibfield{author}{\bibinfo{person}{Xiaojun Chen}, \bibinfo{person}{Ling Ding},
  {and} \bibinfo{person}{Yang Xiang}.} \bibinfo{year}{2021}\natexlab{}.
\newblock \showarticletitle{Neighborhood aggregation based graph attention
  networks for open-world knowledge graph reasoning}.
\newblock \bibinfo{journal}{\emph{Journal of Intelligent \& Fuzzy Systems}}
  (\bibinfo{year}{2021}), \bibinfo{pages}{1--12}.
\newblock


\bibitem[Chen et~al\mbox{.}(2022b)]%
        {FuzzQE}
\bibfield{author}{\bibinfo{person}{Xuelu Chen}, \bibinfo{person}{Ziniu Hu},
  {and} \bibinfo{person}{Yizhou Sun}.} \bibinfo{year}{2022}\natexlab{b}.
\newblock \showarticletitle{Fuzzy Logic Based Logical Query Answering on
  Knowledge Graphs}. In \bibinfo{booktitle}{\emph{Proceedings of the AAAI
  Conference on Artificial Intelligence}}, Vol.~\bibinfo{volume}{36}.
  \bibinfo{pages}{3939--3948}.
\newblock


\bibitem[Chen et~al\mbox{.}(2020)]%
        {chen2020review}
\bibfield{author}{\bibinfo{person}{Xiaojun Chen}, \bibinfo{person}{Shengbin
  Jia}, {and} \bibinfo{person}{Yang Xiang}.} \bibinfo{year}{2020}\natexlab{}.
\newblock \showarticletitle{A review: Knowledge reasoning over knowledge
  graph}.
\newblock \bibinfo{journal}{\emph{Expert Systems with Applications}}
  \bibinfo{volume}{141} (\bibinfo{year}{2020}), \bibinfo{pages}{112948}.
\newblock


\bibitem[Choudhary et~al\mbox{.}(2021)]%
        {PERM}
\bibfield{author}{\bibinfo{person}{Nurendra Choudhary}, \bibinfo{person}{Nikhil
  Rao}, \bibinfo{person}{Sumeet Katariya}, \bibinfo{person}{Karthik Subbian},
  {and} \bibinfo{person}{Chandan Reddy}.} \bibinfo{year}{2021}\natexlab{}.
\newblock \showarticletitle{Probabilistic entity representation model for
  reasoning over knowledge graphs}.
\newblock \bibinfo{journal}{\emph{Advances in Neural Information Processing
  Systems}}  \bibinfo{volume}{34} (\bibinfo{year}{2021}),
  \bibinfo{pages}{23440--23451}.
\newblock


\bibitem[Friedman and Broeck(2020)]%
        {proKGquery}
\bibfield{author}{\bibinfo{person}{Tal Friedman} {and} \bibinfo{person}{Guy
  Broeck}.} \bibinfo{year}{2020}\natexlab{}.
\newblock \showarticletitle{Symbolic querying of vector spaces: Probabilistic
  databases meets relational embeddings}. In
  \bibinfo{booktitle}{\emph{Conference on Uncertainty in Artificial
  Intelligence}}. PMLR, \bibinfo{pages}{1268--1277}.
\newblock


\bibitem[Garg et~al\mbox{.}(2019)]%
        {quantum}
\bibfield{author}{\bibinfo{person}{Dinesh Garg}, \bibinfo{person}{Shajith
  Ikbal}, \bibinfo{person}{Santosh~K Srivastava}, \bibinfo{person}{Harit
  Vishwakarma}, \bibinfo{person}{Hima Karanam}, {and}
  \bibinfo{person}{L~Venkata Subramaniam}.} \bibinfo{year}{2019}\natexlab{}.
\newblock \showarticletitle{Quantum embedding of knowledge for reasoning}.
\newblock \bibinfo{journal}{\emph{Advances in Neural Information Processing
  Systems}}  \bibinfo{volume}{32} (\bibinfo{year}{2019}).
\newblock


\bibitem[Guu et~al\mbox{.}(2015)]%
        {computegraph&multihop}
\bibfield{author}{\bibinfo{person}{Kelvin Guu}, \bibinfo{person}{John Miller},
  {and} \bibinfo{person}{Percy Liang}.} \bibinfo{year}{2015}\natexlab{}.
\newblock \showarticletitle{Traversing knowledge graphs in vector space}.
\newblock \bibinfo{journal}{\emph{arXiv preprint arXiv:1506.01094}}
  (\bibinfo{year}{2015}).
\newblock


\bibitem[Hamaguchi et~al\mbox{.}(2017)]%
        {neib1}
\bibfield{author}{\bibinfo{person}{Takuo Hamaguchi}, \bibinfo{person}{Hidekazu
  Oiwa}, \bibinfo{person}{Masashi Shimbo}, {and} \bibinfo{person}{Yuji
  Matsumoto}.} \bibinfo{year}{2017}\natexlab{}.
\newblock \showarticletitle{Knowledge transfer for out-of-knowledge-base
  entities: A graph neural network approach}.
\newblock \bibinfo{journal}{\emph{arXiv preprint arXiv:1706.05674}}
  (\bibinfo{year}{2017}).
\newblock


\bibitem[Hamilton et~al\mbox{.}(2018)]%
        {GQE}
\bibfield{author}{\bibinfo{person}{Will Hamilton}, \bibinfo{person}{Payal
  Bajaj}, \bibinfo{person}{Marinka Zitnik}, \bibinfo{person}{Dan Jurafsky},
  {and} \bibinfo{person}{Jure Leskovec}.} \bibinfo{year}{2018}\natexlab{}.
\newblock \showarticletitle{Embedding logical queries on knowledge graphs}.
\newblock \bibinfo{journal}{\emph{Advances in neural information processing
  systems}}  \bibinfo{volume}{31} (\bibinfo{year}{2018}).
\newblock


\bibitem[Hamilton et~al\mbox{.}(2017)]%
        {GNN2}
\bibfield{author}{\bibinfo{person}{Will Hamilton}, \bibinfo{person}{Zhitao
  Ying}, {and} \bibinfo{person}{Jure Leskovec}.}
  \bibinfo{year}{2017}\natexlab{}.
\newblock \showarticletitle{Inductive representation learning on large graphs}.
\newblock \bibinfo{journal}{\emph{Advances in neural information processing
  systems}}  \bibinfo{volume}{30} (\bibinfo{year}{2017}).
\newblock


\bibitem[Han et~al\mbox{.}(2020)]%
        {han2020open}
\bibfield{author}{\bibinfo{person}{Jiale Han}, \bibinfo{person}{Bo Cheng},
  {and} \bibinfo{person}{Xu Wang}.} \bibinfo{year}{2020}\natexlab{}.
\newblock \showarticletitle{Open domain question answering based on text
  enhanced knowledge graph with hyperedge infusion}. In
  \bibinfo{booktitle}{\emph{Findings of the Association for Computational
  Linguistics: EMNLP 2020}}. \bibinfo{pages}{1475--1481}.
\newblock


\bibitem[Hartig and Heese(2007)]%
        {FOL1}
\bibfield{author}{\bibinfo{person}{Olaf Hartig} {and} \bibinfo{person}{Ralf
  Heese}.} \bibinfo{year}{2007}\natexlab{}.
\newblock \showarticletitle{The SPARQL query graph model for query
  optimization}. In \bibinfo{booktitle}{\emph{European Semantic Web
  Conference}}. Springer, \bibinfo{pages}{564--578}.
\newblock


\bibitem[Huang et~al\mbox{.}(2022)]%
        {line}
\bibfield{author}{\bibinfo{person}{Zijian Huang}, \bibinfo{person}{Meng-Fen
  Chiang}, {and} \bibinfo{person}{Wang-Chien Lee}.}
  \bibinfo{year}{2022}\natexlab{}.
\newblock \showarticletitle{LinE: Logical Query Reasoning over Hierarchical
  Knowledge Graphs}. In \bibinfo{booktitle}{\emph{Proceedings of the 28th ACM
  SIGKDD Conference on Knowledge Discovery and Data Mining}}.
  \bibinfo{pages}{615--625}.
\newblock


\bibitem[Kapanipathi et~al\mbox{.}(2020)]%
        {fbeate1}
\bibfield{author}{\bibinfo{person}{Pavan Kapanipathi}, \bibinfo{person}{Ibrahim
  Abdelaziz}, \bibinfo{person}{Srinivas Ravishankar}, \bibinfo{person}{Salim
  Roukos}, \bibinfo{person}{Alexander Gray}, \bibinfo{person}{Ramon Astudillo},
  \bibinfo{person}{Maria Chang}, \bibinfo{person}{Cristina Cornelio},
  \bibinfo{person}{Saswati Dana}, \bibinfo{person}{Achille Fokoue},
  {et~al\mbox{.}}} \bibinfo{year}{2020}\natexlab{}.
\newblock \showarticletitle{Leveraging abstract meaning representation for
  knowledge base question answering}.
\newblock \bibinfo{journal}{\emph{arXiv preprint arXiv:2012.01707}}
  (\bibinfo{year}{2020}).
\newblock


\bibitem[Kosasih et~al\mbox{.}(2022)]%
        {kosasih2022towards}
\bibfield{author}{\bibinfo{person}{Edward~Elson Kosasih},
  \bibinfo{person}{Fabrizio Margaroli}, \bibinfo{person}{Simone Gelli},
  \bibinfo{person}{Ajmal Aziz}, \bibinfo{person}{Nick Wildgoose}, {and}
  \bibinfo{person}{Alexandra Brintrup}.} \bibinfo{year}{2022}\natexlab{}.
\newblock \showarticletitle{Towards knowledge graph reasoning for supply chain
  risk management using graph neural networks}.
\newblock \bibinfo{journal}{\emph{International Journal of Production
  Research}} (\bibinfo{year}{2022}), \bibinfo{pages}{1--17}.
\newblock


\bibitem[Krompa{\ss} et~al\mbox{.}(2014)]%
        {probabilistic}
\bibfield{author}{\bibinfo{person}{Denis Krompa{\ss}},
  \bibinfo{person}{Maximilian Nickel}, {and} \bibinfo{person}{Volker Tresp}.}
  \bibinfo{year}{2014}\natexlab{}.
\newblock \showarticletitle{Querying factorized probabilistic triple
  databases}. In \bibinfo{booktitle}{\emph{International Semantic Web
  Conference}}. Springer, \bibinfo{pages}{114--129}.
\newblock


\bibitem[Lao et~al\mbox{.}(2011)]%
        {PRA}
\bibfield{author}{\bibinfo{person}{Ni Lao}, \bibinfo{person}{Tom Mitchell},
  {and} \bibinfo{person}{William Cohen}.} \bibinfo{year}{2011}\natexlab{}.
\newblock \showarticletitle{Random walk inference and learning in a large scale
  knowledge base}. In \bibinfo{booktitle}{\emph{Proceedings of the 2011
  conference on empirical methods in natural language processing}}.
  \bibinfo{pages}{529--539}.
\newblock


\bibitem[Li et~al\mbox{.}(2021)]%
        {li2021memorypath}
\bibfield{author}{\bibinfo{person}{Shuangyin Li}, \bibinfo{person}{Heng Wang},
  \bibinfo{person}{Rong Pan}, {and} \bibinfo{person}{Mingzhi Mao}.}
  \bibinfo{year}{2021}\natexlab{}.
\newblock \showarticletitle{MemoryPath: A deep reinforcement learning framework
  for incorporating memory component into knowledge graph reasoning}.
\newblock \bibinfo{journal}{\emph{Neurocomputing}}  \bibinfo{volume}{419}
  (\bibinfo{year}{2021}), \bibinfo{pages}{273--286}.
\newblock


\bibitem[Liu et~al\mbox{.}(2021)]%
        {liu2021neural}
\bibfield{author}{\bibinfo{person}{Lihui Liu}, \bibinfo{person}{Boxin Du},
  \bibinfo{person}{Heng Ji}, \bibinfo{person}{ChengXiang Zhai}, {and}
  \bibinfo{person}{Hanghang Tong}.} \bibinfo{year}{2021}\natexlab{}.
\newblock \showarticletitle{Neural-Answering Logical Queries on Knowledge
  Graphs}. In \bibinfo{booktitle}{\emph{Proceedings of the 27th ACM SIGKDD
  Conference on Knowledge Discovery \& Data Mining}}.
  \bibinfo{pages}{1087--1097}.
\newblock


\bibitem[Nayyeri et~al\mbox{.}(2021)]%
        {nayyeri2021trans4e}
\bibfield{author}{\bibinfo{person}{Mojtaba Nayyeri},
  \bibinfo{person}{Gokce~Muge Cil}, \bibinfo{person}{Sahar Vahdati},
  \bibinfo{person}{Francesco Osborne}, \bibinfo{person}{Mahfuzur Rahman},
  \bibinfo{person}{Simone Angioni}, \bibinfo{person}{Angelo Salatino},
  \bibinfo{person}{Diego~Reforgiato Recupero}, \bibinfo{person}{Nadezhda
  Vassilyeva}, \bibinfo{person}{Enrico Motta}, {et~al\mbox{.}}}
  \bibinfo{year}{2021}\natexlab{}.
\newblock \showarticletitle{Trans4E: Link prediction on scholarly knowledge
  graphs}.
\newblock \bibinfo{journal}{\emph{Neurocomputing}}  \bibinfo{volume}{461}
  (\bibinfo{year}{2021}), \bibinfo{pages}{530--542}.
\newblock


\bibitem[Reagen et~al\mbox{.}(2016)]%
        {MINERVA}
\bibfield{author}{\bibinfo{person}{Brandon Reagen}, \bibinfo{person}{Paul
  Whatmough}, \bibinfo{person}{Robert Adolf}, \bibinfo{person}{Saketh Rama},
  \bibinfo{person}{Hyunkwang Lee}, \bibinfo{person}{Sae~Kyu Lee},
  \bibinfo{person}{Jos{\'e}~Miguel Hern{\'a}ndez-Lobato},
  \bibinfo{person}{Gu-Yeon Wei}, {and} \bibinfo{person}{David Brooks}.}
  \bibinfo{year}{2016}\natexlab{}.
\newblock \showarticletitle{Minerva: Enabling low-power, highly-accurate deep
  neural network accelerators}. In \bibinfo{booktitle}{\emph{2016 ACM/IEEE 43rd
  Annual International Symposium on Computer Architecture (ISCA)}}. IEEE,
  \bibinfo{pages}{267--278}.
\newblock


\bibitem[Ren et~al\mbox{.}(2021)]%
        {lego}
\bibfield{author}{\bibinfo{person}{Hongyu Ren}, \bibinfo{person}{Hanjun Dai},
  \bibinfo{person}{Bo Dai}, \bibinfo{person}{Xinyun Chen},
  \bibinfo{person}{Michihiro Yasunaga}, \bibinfo{person}{Haitian Sun},
  \bibinfo{person}{Dale Schuurmans}, \bibinfo{person}{Jure Leskovec}, {and}
  \bibinfo{person}{Denny Zhou}.} \bibinfo{year}{2021}\natexlab{}.
\newblock \showarticletitle{Lego: Latent execution-guided reasoning for
  multi-hop question answering on knowledge graphs}. In
  \bibinfo{booktitle}{\emph{International Conference on Machine Learning}}.
  PMLR, \bibinfo{pages}{8959--8970}.
\newblock


\bibitem[Ren et~al\mbox{.}(2020)]%
        {Q2B}
\bibfield{author}{\bibinfo{person}{Hongyu Ren}, \bibinfo{person}{Weihua Hu},
  {and} \bibinfo{person}{Jure Leskovec}.} \bibinfo{year}{2020}\natexlab{}.
\newblock \showarticletitle{Query2box: Reasoning over knowledge graphs in
  vector space using box embeddings}.
\newblock \bibinfo{journal}{\emph{arXiv preprint arXiv:2002.05969}}
  (\bibinfo{year}{2020}).
\newblock


\bibitem[Ren and Leskovec(2020)]%
        {BetaE&KGreasoning}
\bibfield{author}{\bibinfo{person}{Hongyu Ren} {and} \bibinfo{person}{Jure
  Leskovec}.} \bibinfo{year}{2020}\natexlab{}.
\newblock \showarticletitle{Beta embeddings for multi-hop logical reasoning in
  knowledge graphs}.
\newblock \bibinfo{journal}{\emph{Advances in Neural Information Processing
  Systems}}  \bibinfo{volume}{33} (\bibinfo{year}{2020}),
  \bibinfo{pages}{19716--19726}.
\newblock


\bibitem[Rossi et~al\mbox{.}(2021)]%
        {rossi2021knowledge}
\bibfield{author}{\bibinfo{person}{Andrea Rossi}, \bibinfo{person}{Denilson
  Barbosa}, \bibinfo{person}{Donatella Firmani}, \bibinfo{person}{Antonio
  Matinata}, {and} \bibinfo{person}{Paolo Merialdo}.}
  \bibinfo{year}{2021}\natexlab{}.
\newblock \showarticletitle{Knowledge graph embedding for link prediction: A
  comparative analysis}.
\newblock \bibinfo{journal}{\emph{ACM Transactions on Knowledge Discovery from
  Data (TKDD)}} \bibinfo{volume}{15}, \bibinfo{number}{2}
  (\bibinfo{year}{2021}), \bibinfo{pages}{1--49}.
\newblock


\bibitem[Sadeghian et~al\mbox{.}(2021)]%
        {emb5}
\bibfield{author}{\bibinfo{person}{Ali Sadeghian},
  \bibinfo{person}{Mohammadreza Armandpour}, \bibinfo{person}{Anthony Colas},
  {and} \bibinfo{person}{Daisy~Zhe Wang}.} \bibinfo{year}{2021}\natexlab{}.
\newblock \showarticletitle{ChronoR: rotation based temporal knowledge graph
  embedding}. In \bibinfo{booktitle}{\emph{Proceedings of the AAAI Conference
  on Artificial Intelligence}}, Vol.~\bibinfo{volume}{35}.
  \bibinfo{pages}{6471--6479}.
\newblock


\bibitem[Sahlgren(2008)]%
        {distributional}
\bibfield{author}{\bibinfo{person}{Magnus Sahlgren}.}
  \bibinfo{year}{2008}\natexlab{}.
\newblock \showarticletitle{The distributional hypothesis}.
\newblock \bibinfo{journal}{\emph{Italian Journal of Disability Studies}}
  \bibinfo{volume}{20} (\bibinfo{year}{2008}), \bibinfo{pages}{33--53}.
\newblock


\bibitem[Schlichtkrull et~al\mbox{.}(2018)]%
        {R-GCN}
\bibfield{author}{\bibinfo{person}{Michael Schlichtkrull},
  \bibinfo{person}{Thomas~N Kipf}, \bibinfo{person}{Peter Bloem},
  \bibinfo{person}{Rianne van~den Berg}, \bibinfo{person}{Ivan Titov}, {and}
  \bibinfo{person}{Max Welling}.} \bibinfo{year}{2018}\natexlab{}.
\newblock \showarticletitle{Modeling relational data with graph convolutional
  networks}. In \bibinfo{booktitle}{\emph{European semantic web conference}}.
  Springer, \bibinfo{pages}{593--607}.
\newblock


\bibitem[Shan et~al\mbox{.}(2018)]%
        {shan2018confidence}
\bibfield{author}{\bibinfo{person}{Yingchun Shan}, \bibinfo{person}{Chenyang
  Bu}, \bibinfo{person}{Xiaojian Liu}, \bibinfo{person}{Shengwei Ji}, {and}
  \bibinfo{person}{Lei Li}.} \bibinfo{year}{2018}\natexlab{}.
\newblock \showarticletitle{Confidence-aware negative sampling method for noisy
  knowledge graph embedding}. In \bibinfo{booktitle}{\emph{2018 IEEE
  International Conference on Big Knowledge (ICBK)}}. IEEE,
  \bibinfo{pages}{33--40}.
\newblock


\bibitem[Shang et~al\mbox{.}(2019a)]%
        {GNN1}
\bibfield{author}{\bibinfo{person}{Chao Shang}, \bibinfo{person}{Yun Tang},
  \bibinfo{person}{Jing Huang}, \bibinfo{person}{Jinbo Bi},
  \bibinfo{person}{Xiaodong He}, {and} \bibinfo{person}{Bowen Zhou}.}
  \bibinfo{year}{2019}\natexlab{a}.
\newblock \showarticletitle{End-to-end structure-aware convolutional networks
  for knowledge base completion}. In \bibinfo{booktitle}{\emph{Proceedings of
  the AAAI Conference on Artificial Intelligence}}.
  \bibinfo{pages}{3060--3067}.
\newblock


\bibitem[Shang et~al\mbox{.}(2019b)]%
        {ConvE}
\bibfield{author}{\bibinfo{person}{Chao Shang}, \bibinfo{person}{Yun Tang},
  \bibinfo{person}{Jing Huang}, \bibinfo{person}{Jinbo Bi},
  \bibinfo{person}{Xiaodong He}, {and} \bibinfo{person}{Bowen Zhou}.}
  \bibinfo{year}{2019}\natexlab{b}.
\newblock \showarticletitle{End-to-end structure-aware convolutional networks
  for knowledge base completion}. In \bibinfo{booktitle}{\emph{Proceedings of
  the AAAI Conference on Artificial Intelligence}}.
  \bibinfo{pages}{3060--3067}.
\newblock


\bibitem[Shao et~al\mbox{.}(2021)]%
        {shao2021dskrl}
\bibfield{author}{\bibinfo{person}{Tianyang Shao}, \bibinfo{person}{Xinyi Li},
  \bibinfo{person}{Xiang Zhao}, \bibinfo{person}{Hao Xu}, {and}
  \bibinfo{person}{Weidong Xiao}.} \bibinfo{year}{2021}\natexlab{}.
\newblock \showarticletitle{DSKRL: A dissimilarity-support-aware knowledge
  representation learning framework on noisy knowledge graph}.
\newblock \bibinfo{journal}{\emph{Neurocomputing}}  \bibinfo{volume}{461}
  (\bibinfo{year}{2021}), \bibinfo{pages}{608--617}.
\newblock


\bibitem[Shi and Weninger(2018)]%
        {opencomp}
\bibfield{author}{\bibinfo{person}{Baoxu Shi} {and} \bibinfo{person}{Tim
  Weninger}.} \bibinfo{year}{2018}\natexlab{}.
\newblock \showarticletitle{Open-world knowledge graph completion}. In
  \bibinfo{booktitle}{\emph{Proceedings of the AAAI conference on artificial
  intelligence}}, Vol.~\bibinfo{volume}{32}.
\newblock


\bibitem[Sun et~al\mbox{.}(2020a)]%
        {faithful}
\bibfield{author}{\bibinfo{person}{Haitian Sun}, \bibinfo{person}{Andrew
  Arnold}, \bibinfo{person}{Tania Bedrax~Weiss}, \bibinfo{person}{Fernando
  Pereira}, {and} \bibinfo{person}{William~W Cohen}.}
  \bibinfo{year}{2020}\natexlab{a}.
\newblock \showarticletitle{Faithful embeddings for knowledge base queries}.
\newblock \bibinfo{journal}{\emph{Advances in Neural Information Processing
  Systems}}  \bibinfo{volume}{33} (\bibinfo{year}{2020}),
  \bibinfo{pages}{22505--22516}.
\newblock


\bibitem[Sun et~al\mbox{.}(2019)]%
        {RotatE}
\bibfield{author}{\bibinfo{person}{Zhiqing Sun}, \bibinfo{person}{Zhi-Hong
  Deng}, \bibinfo{person}{Jian-Yun Nie}, {and} \bibinfo{person}{Jian Tang}.}
  \bibinfo{year}{2019}\natexlab{}.
\newblock \showarticletitle{Rotate: Knowledge graph embedding by relational
  rotation in complex space}.
\newblock \bibinfo{journal}{\emph{arXiv preprint arXiv:1902.10197}}
  (\bibinfo{year}{2019}).
\newblock


\bibitem[Sun et~al\mbox{.}(2020b)]%
        {neib2}
\bibfield{author}{\bibinfo{person}{Zequn Sun}, \bibinfo{person}{Chengming
  Wang}, \bibinfo{person}{Wei Hu}, \bibinfo{person}{Muhao Chen},
  \bibinfo{person}{Jian Dai}, \bibinfo{person}{Wei Zhang}, {and}
  \bibinfo{person}{Yuzhong Qu}.} \bibinfo{year}{2020}\natexlab{b}.
\newblock \showarticletitle{Knowledge graph alignment network with gated
  multi-hop neighborhood aggregation}. In \bibinfo{booktitle}{\emph{Proceedings
  of the AAAI Conference on Artificial Intelligence}}.
  \bibinfo{pages}{222--229}.
\newblock


\bibitem[Tang et~al\mbox{.}(2022)]%
        {ABIN}
\bibfield{author}{\bibinfo{person}{Zhenwei Tang}, \bibinfo{person}{Shichao
  Pei}, \bibinfo{person}{Xi Peng}, \bibinfo{person}{Fuzhen Zhuang},
  \bibinfo{person}{Xiangliang Zhang}, {and} \bibinfo{person}{Robert
  Hoehndorf}.} \bibinfo{year}{2022}\natexlab{}.
\newblock \showarticletitle{Joint Abductive and Inductive Neural Logical
  Reasoning}.
\newblock \bibinfo{journal}{\emph{arXiv preprint arXiv:2205.14591}}
  (\bibinfo{year}{2022}).
\newblock


\bibitem[Tiwari et~al\mbox{.}(2021)]%
        {dapath}
\bibfield{author}{\bibinfo{person}{Prayag Tiwari}, \bibinfo{person}{Hongyin
  Zhu}, {and} \bibinfo{person}{Hari~Mohan Pandey}.}
  \bibinfo{year}{2021}\natexlab{}.
\newblock \showarticletitle{DAPath: Distance-aware knowledge graph reasoning
  based on deep reinforcement learning}.
\newblock \bibinfo{journal}{\emph{Neural Networks}}  \bibinfo{volume}{135}
  (\bibinfo{year}{2021}), \bibinfo{pages}{1--12}.
\newblock


\bibitem[Toutanova and Chen(2015)]%
        {FB15k-237}
\bibfield{author}{\bibinfo{person}{Kristina Toutanova} {and}
  \bibinfo{person}{Danqi Chen}.} \bibinfo{year}{2015}\natexlab{}.
\newblock \showarticletitle{Observed versus latent features for knowledge base
  and text inference}. In \bibinfo{booktitle}{\emph{Proceedings of the 3rd
  workshop on continuous vector space models and their compositionality}}.
  \bibinfo{pages}{57--66}.
\newblock


\bibitem[Trouillon et~al\mbox{.}(2016)]%
        {ComplEx}
\bibfield{author}{\bibinfo{person}{Th{\'e}o Trouillon},
  \bibinfo{person}{Johannes Welbl}, \bibinfo{person}{Sebastian Riedel},
  \bibinfo{person}{{\'E}ric Gaussier}, {and} \bibinfo{person}{Guillaume
  Bouchard}.} \bibinfo{year}{2016}\natexlab{}.
\newblock \showarticletitle{Complex embeddings for simple link prediction}. In
  \bibinfo{booktitle}{\emph{International conference on machine learning}}.
  PMLR, \bibinfo{pages}{2071--2080}.
\newblock


\bibitem[Vashishth et~al\mbox{.}(2020)]%
        {emb3}
\bibfield{author}{\bibinfo{person}{Shikhar Vashishth}, \bibinfo{person}{Soumya
  Sanyal}, \bibinfo{person}{Vikram Nitin}, \bibinfo{person}{Nilesh Agrawal},
  {and} \bibinfo{person}{Partha Talukdar}.} \bibinfo{year}{2020}\natexlab{}.
\newblock \showarticletitle{Interacte: Improving convolution-based knowledge
  graph embeddings by increasing feature interactions}. In
  \bibinfo{booktitle}{\emph{Proceedings of the AAAI conference on artificial
  intelligence}}, Vol.~\bibinfo{volume}{34}. \bibinfo{pages}{3009--3016}.
\newblock


\bibitem[Vilnis et~al\mbox{.}(2018)]%
        {vilnis2018probabilistic}
\bibfield{author}{\bibinfo{person}{Luke Vilnis}, \bibinfo{person}{Xiang Li},
  \bibinfo{person}{Shikhar Murty}, {and} \bibinfo{person}{Andrew McCallum}.}
  \bibinfo{year}{2018}\natexlab{}.
\newblock \showarticletitle{Probabilistic embedding of knowledge graphs with
  box lattice measures}.
\newblock \bibinfo{journal}{\emph{arXiv preprint arXiv:1805.06627}}
  (\bibinfo{year}{2018}).
\newblock


\bibitem[Wan et~al\mbox{.}(2020)]%
        {wan2020adaptive}
\bibfield{author}{\bibinfo{person}{Guojia Wan}, \bibinfo{person}{Bo Du},
  \bibinfo{person}{Shirui Pan}, {and} \bibinfo{person}{Jia Wu}.}
  \bibinfo{year}{2020}\natexlab{}.
\newblock \showarticletitle{Adaptive knowledge subgraph ensemble for robust and
  trustworthy knowledge graph completion}.
\newblock \bibinfo{journal}{\emph{World Wide Web}} \bibinfo{volume}{23},
  \bibinfo{number}{1} (\bibinfo{year}{2020}), \bibinfo{pages}{471--490}.
\newblock


\bibitem[Wan et~al\mbox{.}(2021)]%
        {wan2021reasoning}
\bibfield{author}{\bibinfo{person}{Guojia Wan}, \bibinfo{person}{Shirui Pan},
  \bibinfo{person}{Chen Gong}, \bibinfo{person}{Chuan Zhou}, {and}
  \bibinfo{person}{Gholamreza Haffari}.} \bibinfo{year}{2021}\natexlab{}.
\newblock \showarticletitle{Reasoning like human: Hierarchical reinforcement
  learning for knowledge graph reasoning}. In
  \bibinfo{booktitle}{\emph{Proceedings of the Twenty-Ninth International
  Conference on International Joint Conferences on Artificial Intelligence}}.
  \bibinfo{pages}{1926--1932}.
\newblock


\bibitem[Wang et~al\mbox{.}(2020)]%
        {wang2020adrl}
\bibfield{author}{\bibinfo{person}{Qi Wang}, \bibinfo{person}{Yongsheng Hao},
  {and} \bibinfo{person}{Jie Cao}.} \bibinfo{year}{2020}\natexlab{}.
\newblock \showarticletitle{ADRL: An attention-based deep reinforcement
  learning framework for knowledge graph reasoning}.
\newblock \bibinfo{journal}{\emph{Knowledge-Based Systems}}
  \bibinfo{volume}{197} (\bibinfo{year}{2020}), \bibinfo{pages}{105910}.
\newblock


\bibitem[Xiong et~al\mbox{.}(2017)]%
        {NELL995}
\bibfield{author}{\bibinfo{person}{Wenhan Xiong}, \bibinfo{person}{Thien
  Hoang}, {and} \bibinfo{person}{William~Yang Wang}.}
  \bibinfo{year}{2017}\natexlab{}.
\newblock \showarticletitle{Deeppath: A reinforcement learning method for
  knowledge graph reasoning}.
\newblock \bibinfo{journal}{\emph{arXiv preprint arXiv:1707.06690}}
  (\bibinfo{year}{2017}).
\newblock


\bibitem[Ye et~al\mbox{.}(2022)]%
        {ye2022comprehensive}
\bibfield{author}{\bibinfo{person}{Zi Ye}, \bibinfo{person}{Yogan~Jaya Kumar},
  \bibinfo{person}{Goh~Ong Sing}, \bibinfo{person}{Fengyan Song}, {and}
  \bibinfo{person}{Junsong Wang}.} \bibinfo{year}{2022}\natexlab{}.
\newblock \showarticletitle{A Comprehensive Survey of Graph Neural Networks for
  Knowledge Graphs}.
\newblock \bibinfo{journal}{\emph{IEEE Access}}  \bibinfo{volume}{10}
  (\bibinfo{year}{2022}), \bibinfo{pages}{75729--75741}.
\newblock


\bibitem[Yu et~al\mbox{.}(2021)]%
        {Kg-fid}
\bibfield{author}{\bibinfo{person}{Donghan Yu}, \bibinfo{person}{Chenguang
  Zhu}, \bibinfo{person}{Yuwei Fang}, \bibinfo{person}{Wenhao Yu},
  \bibinfo{person}{Shuohang Wang}, \bibinfo{person}{Yichong Xu},
  \bibinfo{person}{Xiang Ren}, \bibinfo{person}{Yiming Yang}, {and}
  \bibinfo{person}{Michael Zeng}.} \bibinfo{year}{2021}\natexlab{}.
\newblock \showarticletitle{Kg-fid: Infusing knowledge graph in
  fusion-in-decoder for open-domain question answering}.
\newblock \bibinfo{journal}{\emph{arXiv preprint arXiv:2110.04330}}
  (\bibinfo{year}{2021}).
\newblock


\bibitem[Zhang et~al\mbox{.}(2019)]%
        {emb1}
\bibfield{author}{\bibinfo{person}{Shuai Zhang}, \bibinfo{person}{Yi Tay},
  \bibinfo{person}{Lina Yao}, {and} \bibinfo{person}{Qi Liu}.}
  \bibinfo{year}{2019}\natexlab{}.
\newblock \showarticletitle{Quaternion knowledge graph embeddings}.
\newblock \bibinfo{journal}{\emph{Advances in neural information processing
  systems}}  \bibinfo{volume}{32} (\bibinfo{year}{2019}).
\newblock


\bibitem[Zhang et~al\mbox{.}(2022)]%
        {zhang2022knowledge}
\bibfield{author}{\bibinfo{person}{Wen Zhang}, \bibinfo{person}{Jiaoyan Chen},
  \bibinfo{person}{Juan Li}, \bibinfo{person}{Zezhong Xu},
  \bibinfo{person}{Jeff~Z Pan}, {and} \bibinfo{person}{Huajun Chen}.}
  \bibinfo{year}{2022}\natexlab{}.
\newblock \showarticletitle{Knowledge Graph Reasoning with Logics and
  Embeddings: Survey and Perspective}.
\newblock \bibinfo{journal}{\emph{arXiv preprint arXiv:2202.07412}}
  (\bibinfo{year}{2022}).
\newblock


\bibitem[Zhang et~al\mbox{.}(2021a)]%
        {zhang2021fact}
\bibfield{author}{\bibinfo{person}{Yao Zhang}, \bibinfo{person}{Peiyao Li},
  \bibinfo{person}{Hongru Liang}, \bibinfo{person}{Adam Jatowt}, {and}
  \bibinfo{person}{Zhenglu Yang}.} \bibinfo{year}{2021}\natexlab{a}.
\newblock \showarticletitle{Fact-Tree Reasoning for N-ary Question Answering
  over Knowledge Graphs}.
\newblock \bibinfo{journal}{\emph{arXiv preprint arXiv:2108.08297}}
  (\bibinfo{year}{2021}).
\newblock


\bibitem[Zhang and Yao(2022)]%
        {redGCNneib2}
\bibfield{author}{\bibinfo{person}{Yongqi Zhang} {and}
  \bibinfo{person}{Quanming Yao}.} \bibinfo{year}{2022}\natexlab{}.
\newblock \showarticletitle{Knowledge graph reasoning with relational digraph}.
  In \bibinfo{booktitle}{\emph{Proceedings of the ACM Web Conference 2022}}.
  \bibinfo{pages}{912--924}.
\newblock


\bibitem[Zhang et~al\mbox{.}(2021b)]%
        {ConE}
\bibfield{author}{\bibinfo{person}{Zhanqiu Zhang}, \bibinfo{person}{Jie Wang},
  \bibinfo{person}{Jiajun Chen}, \bibinfo{person}{Shuiwang Ji}, {and}
  \bibinfo{person}{Feng Wu}.} \bibinfo{year}{2021}\natexlab{b}.
\newblock \showarticletitle{Cone: Cone embeddings for multi-hop reasoning over
  knowledge graphs}.
\newblock \bibinfo{journal}{\emph{Advances in Neural Information Processing
  Systems}}  \bibinfo{volume}{34} (\bibinfo{year}{2021}),
  \bibinfo{pages}{19172--19183}.
\newblock


\bibitem[Zhou et~al\mbox{.}(2022)]%
        {JointE}
\bibfield{author}{\bibinfo{person}{Zhehui Zhou}, \bibinfo{person}{Can Wang},
  \bibinfo{person}{Yan Feng}, {and} \bibinfo{person}{Defang Chen}.}
  \bibinfo{year}{2022}\natexlab{}.
\newblock \showarticletitle{JointE: Jointly utilizing 1D and 2D convolution for
  knowledge graph embedding}.
\newblock \bibinfo{journal}{\emph{Knowledge-Based Systems}}
  \bibinfo{volume}{240} (\bibinfo{year}{2022}), \bibinfo{pages}{108100}.
\newblock


\bibitem[Zou et~al\mbox{.}(2011)]%
        {FOL2}
\bibfield{author}{\bibinfo{person}{Lei Zou}, \bibinfo{person}{Jinghui Mo},
  \bibinfo{person}{Lei Chen}, \bibinfo{person}{M~Tamer {\"O}zsu}, {and}
  \bibinfo{person}{Dongyan Zhao}.} \bibinfo{year}{2011}\natexlab{}.
\newblock \showarticletitle{gStore: answering SPARQL queries via subgraph
  matching}.
\newblock \bibinfo{journal}{\emph{Proceedings of the VLDB Endowment}}
  \bibinfo{volume}{4}, \bibinfo{number}{8} (\bibinfo{year}{2011}),
  \bibinfo{pages}{482--493}.
\newblock


\end{thebibliography}
